\documentclass[preprint,12pt]{elsarticle}

\usepackage{amssymb}
\usepackage{amsmath}
\usepackage[T1]{fontenc}
\usepackage{color}
\usepackage{array}
\usepackage{mathtools}
\usepackage{graphicx}
\usepackage{algorithm}
\usepackage{algpseudocode}
\usepackage{multirow}
\usepackage{adjustbox}
\usepackage{soul,subcaption}
\usepackage{url}
\usepackage[inline]{enumitem}
\journal{arXiv}

\begin{document}

\begin{frontmatter}

\author[1]{Hassan Eshkiki}\ead{h.g.eshkiki@swansea.ac.uk}
\author[1]{Sarah Costa}\ead{2038107@swansea.ac.uk}
\author[2]{Mostafa Mohammadpour}\ead{mohammadpour@gtec.at}
\author[4]{Farinaz Tanhaei}\ead{farinaztanhaiiedu@gmail.com}
\author[4]{Christopher H. George}\ead{christopher.george@swansea.ac.uk}
\author[1]{Fabio Caraffini\corref{corr}}\ead{fabio.caraffini@swansea.ac.uk}\ead[url]{https://orcid.org/0000-0001-9199-7368}

\address[1]{Department of Computer Science, Swansea University, Swansea SA1 8EN, UK}
\address[2]{Department of Computational Perception, Johannes Kepler University, Linz, Austria}
\address[4]{Swansea University Medical School, Faculty of Medicine, Health and Life Sciences, Swansea SA2 8PP, UK}

\cortext[corr]{Corresponding author}
 \title{Multi-Temporal Frames Projection for Dynamic Processes Fusion in Fluorescence Microscopy}

\begin{abstract}
Fluorescence microscopy is widely employed for the analysis of living biological samples; however, the utility of the resulting recordings is frequently constrained by noise, temporal variability, and inconsistent visualisation of signals that oscillate over time. We present a unique computational framework that integrates information from multiple time-resolved frames into a single high-quality image, while preserving the underlying biological content of the original video. We evaluate the proposed method through an extensive number of configurations (n = 111) and on a challenging dataset comprising dynamic, heterogeneous, and morphologically complex 2D monolayers of cardiac cells. Results show that our framework, which consists of a combination of explainable techniques from different computer vision application fields, is capable of generating composite images that preserve and enhance the quality and information of individual microscopy frames, yielding $44\%$ average increase in cell count compared to previous methods. The proposed pipeline is applicable to other imaging domains that require the fusion of multi-temporal image stacks into high-quality 2D images, thereby facilitating annotation and downstream segmentation.
\end{abstract}


\end{frontmatter}

\section{Introduction}\label{sec:Intro}

Fluorescence microscopy (FM) is an established technique for visualising dynamic processes in living cells and organisms with high spatial and temporal resolution. The availability of fluorescent probes with distinct spectral properties makes this method highly versatile, enabling researchers to observe cellular behaviour and molecular components, including proteins, nucleic acids, and signalling molecules, like calcium ions ($Ca^{2+}$), in real time \cite{nicoli2025synthetic}. Despite its widespread use, FM faces several technical challenges. Photobleaching and phototoxicity remain significant limitations, as prolonged exposure to excitation light can degrade fluorophores and damage living samples, potentially introducing artefacts and reducing image quality \cite{icha2017phototoxicity}. 

To address these issues, advanced imaging modalities, such as Confocal Laser Scanning Microscopy (CLSM), have been developed. CLSM improves optical sectioning by capturing images in discrete focal planes within the biological sample, which reduces out-of-focus light and enables the accurate reconstruction of its structure \cite{pawley2006handbook}. This is particularly valuable for studying subcellular organisation and live cellular processes, including changes in whole-cell morphology and contractility \cite{burbaum2021molecular}, and intracellular dynamics, such as protein-protein interactions and molecular signalling mechanisms \cite{gilbert2020calcium} in living cells and tissue models. However, like conventional FM, CLSM  is affected by the prolonged exposure of biological samples to focused laser beams, which limits the permissible imaging time to avoid damaging living samples, at the expenses of image quality \cite{icha2017phototoxicity}. Other limitations of CLSM include optical scattering, which causes background noise \cite{herron2012optical}, motion and focus shifts, which blur object boundaries \cite{presotto2022long}, and illumination variability, that leads to inconsistent contrast \cite{Kato2024}.

Analysing high-dimensional data from live-cell CLSM experiments remains challenging, especially in tissue models that contain densely packed, heterogeneous, and aggregated cells with undefined boundaries \cite{caraffini2024towards,bib:costa2025AIiH}. Extracting quantitative information from these models requires segmenting individual cellular structures within the acquired images. However, image-quality issues, which are common in microscopy \cite{Kato2024}, and the complexity of cellular networks often degrade the performance of existing segmentation algorithms, like \cite{Cellpose2021}, which are usually trained on static, high-contrast datasets acquired under controlled imaging conditions and containing mainly cells characterised by a high degree of sphericity and distinct edges. When presented with real-world data, these models cause oversegmentation \cite{caraffini2024towards}, or fail to detect cells altogether \cite{Fang2023}, producing inaccurate measurements and driving flawed biological conclusions. 

Without an accurate system for automated cell segmentation, CLSM recordings are often segmented manually using a frame-by-frame approach. This process, other than being extremely time-consuming, is highly prone to segmentation errors, particularly when the Fluorescence Intensity (FI) of cells within the recordings oscillates over time, as in the case of $Ca^{2+}$ signalling experiments \cite{herron2012optical}, making cells inconsistently visible in each frame, or when subtle, low-intensity structural features are obscured by noise.

To address these challenges, we have developed a systematic workflow designed to generate preprocessing pipelines that enable the fusion of multi-temporal frame sequences acquired from FM recordings into high-quality composite images for downstream segmentation and analysis. We evaluate our approach
using a challenging dataset of CLSM $Ca^{2+}$ signalling experiments performed on dynamic, heterogeneous and morphologically complex cardiac cell networks, thereby constituting an exemplar case study. By using a combination of z-projection and image enhancement methods, our pipeline aims at effectively resolving issues related to noise, contrast, temporal dynamicity and inconsistent visibility, allowing the visualisation of oscillatory signals occurring at different temporal intervals within these cells into a comprehensive two-dimensional (2D) representation of the living sample. By doing this, we seek to bridge the gap between raw FM data acquisition and their exploitation for meaningful biological analysis.


\section{Related work}\label{sec:background}
\subsection{Classical Preprocessing Methods in Microscopy}\label{sec:classicalpreproc}
 
Preprocessing plays a pivotal role in the preparation of imaging data for downstream segmentation and analysis \cite{Fang2023}. Numerous methodologies, ranging from traditional filtering techniques to advanced Computer Vision (CV) methods integrated with Deep Learning (DL) pipelines \cite{agarwal2024analysis, bindal2022systematic}, have been devised to address the challenges inherent in raw microscopy data. The overarching goal of these approaches is to enhance any suboptimal quality of images, thus mitigating the risk of poor segmentation accuracy, especially in semantic (i.e., foreground vs background) segmentation tasks \cite{Kato2024}.

Classical image enhancement methods, including Histogram Equalisation (HE) \cite{agarwal2024analysis}, Gamma Correction (GC) \cite{cao2018contrast}, and noise filtering techniques \cite{bindal2022systematic}, remain fundamental and widely used approaches. HE, for example, has been shown to effectively improve foreground-to-background contrast in ophthalmoscopy images, enhancing the visualisation of vascular microstructures at the back of the eye and demonstrating the improved diagnostic potential of optical images when the appropriate preprocessing methods are applied  \cite{ningsih2020improving}. 

In cardiac cell imaging, the application of contrast stretching in conjunction with Gaussian Filtering (GF) has been proven to enhance the segmentation of low-contrast electron microscopy scans of cardiomyocyte (CM) z-disks \cite{Khadangi2019}. These are nanoscopic structures involved in cardiac contraction, often difficult to visualise even with sophisticated super-resolution technologies. Z-disks segmentation particularly improves when probability-weighted mean curves are used to inform histogram-based thresholding, providing a systematic approach to detect low-contrast structures within high-resolution but noisy microscopy datasets \cite{Khadangi2019}. The application of GF for histograms smoothing has also been associated with the effective suppression of intensity noise in FM images \cite{Fang2023}. Furthermore, when coupled with Rosin’s thresholding and ellipse fitting, this approach successfully separated overlapping cells in segmentation tasks \cite{Fang2023}, highlighting the continued utility of histogram-based and noise filtering methods in handling biological complexity. 

\subsection{Hardware-Based Tools}\label{sec:hardware}

Alongside traditional methods, hardware-based and interactive tools have emerged as powerful means of improving microscopy data quality during both acquisition and annotation. ImageJ \cite{schneider2012nih} and Fiji \cite{schindelin2012fiji} are widely used software for the post-acquisition processing of multi-dimensional microscopy datasets. Despite the extensive availability of contrast enhancement, de-noising and segmentation plugins, these platforms were designed as general-scope tools for scientific image processing rather than tailored solutions for microscopy. Several studies have therefore highlighted the suboptimal performance of these software on complex microscopy data, yielding image distortion and inaccurate segmentation \cite{ASH2021108416}. 

To overcome these challenges, specialised software has been designed for microscopy images. Interactive cell segmentation tools, such as Ilastik, have been used in combination with powerful microscopy hardware, like Airyscan confocal microscopes, to robustly handle variability in cellular image brightness and staining, achieving improved image quality directly from the acquisition stage \cite{ibrahim2024}. 
 
By performing a portion of the processing at the hardware level, this approach facilitates the automation of downstream processing and segmentation tasks using interactive tools \cite{ibrahim2024}. A similar approach was adopted in \cite{Salvi2019}, where processing methods were exploited within an automated pipeline for results generation, thereby improving the segmentation of three-dimensional (3D) cellular clusters. These findings demonstrate the importance of image preparation for visual quality enhancement, but also for the refinement and automation of quantitative data extraction from challenging datasets.

\subsection{Deep-Learning Approaches}\label{sec:DLapproach}
The layered architecture of DL algorithms enables the integration of image processing directly within automated segmentation pipelines. For example, \cite{Kato2024} proposed a two-stage Neural Network (NN) within which intermediate layers generate feature maps that enhance low-quality input images, producing sharper object boundaries and reducing noise. These maps serve as improved input for the subsequent segmentation stage, also implemented within the DL pipeline \cite{Kato2024}. These DL-driven methods can additionally incorporate denoising and contrast enhancement techniques directly into their learned representations, reducing their reliance on handcrafted processing.

DL-based frameworks resolve the limitations associated with the use of general-purpose software for microscopy image processing and analysis (Section \ref{sec:hardware}) by enabling the training of optimised architectures on specific datasets. However, model training requires the manual annotation of high-quality and large imaging datasets, as the segmentation performance of DL algorithms is highly impacted by the number and quality of the training and input data. Therefore, even in DL, classic processing approaches remain useful. This synergy between image processing and modelling is highlighted in \cite{leyendecker22a}, which shows how image normalisation and dataset partitioning can be systematically combined with hyperparameter tuning to optimise cell type classification using DL.

\subsection{Multi-Dimensional Image Fusion for Video-to-Image  Integration}\label{sec:Imagefusion}

Image fusion allows the integration of multi-dimensional data to generate composite 2D representations of digitised scenes or objects of interest. 2D images obtained through fusion display a higher resolution and contain more detailed information than the original data used for their generation \cite{singh2023review}. Image fusion methods have been extensively evaluated for the integration of medical scans from different imaging modalities (e.g., CT, MRI, and PET), to produce fused outputs with improved diagnostic potential \cite{singh2023review}. In microscopy, image fusion has been employed to combine multi-modal data \cite{van2015image}, acquired by imaging the sample of interest using different microscopy modalities, multi-focus images \cite{cheng2024multi}, generated by fusing out-of-focus and in-focus microscopy images for improved quality, and multi-volumetric data, to visualise the architecture of 3D models in 2D, including 3D cellular networks \cite{zhou2025universal}. To the best of our knowledge, the use of these techniques for the fusion of multi-temporal data from live FM experiments on fluorescent samples where the intensity of the signal varies over time (i.e., oscillatory $Ca^{2+}$ signals) has not yet been evaluated. 

Amongst all image fusion techniques, Z-projections provide a complementary pathway to integrate live imaging data into 2D visualisations, while improving downstream segmentation accuracy \cite{chao2021new}. Techniques such as Maximum Intensity Projection (MIP) have been successfully incorporated into segmentation pipelines for PET and CT scans, improving lesion detection over purely volumetric U-Net models \cite{Constantino2025}. Working with 2D MIP images has also been shown to substantially reduce annotation times, facilitating the manual generation of Ground Truth (GT) labels for DL model training. In \cite{guo20243}, for example, annotations from 2D MIP images were used to generate pseudo-GT labels to train a 3D vascular segmentation network. Other techniques, including minimum intensity projection and local MIP \cite{cheng2024multi}, are also commonly employed. However, these methods often fail to consider variability within image stacks, generating composite representations that provide incomplete approximations of the original recordings. Furthermore, z-projections are rarely integrated with image enhancement methods and frequently require additional processing for downstream analysis. We address these limitations by proposing a unified framework that combines z-projection methods sensitive to inter-stack variability with image enhancement strategies to improve image fusion, quality and annotation accuracy in temporally-resolved imaging datasets.

\section{General Framework}\label{sec:generalFramework}

\subsection{Aims and Objectives}\label{sec:aims}
This study devises a novel methodological framework for multi-temporal image fusion that integrates information from FM recordings into composite 2D images, while simultaneously improving image quality. By fulfilling this aim, we address a problem formulated as the superposition of two main challenges. The first arises from the low quality of FM data, which are often affected by noise and imaging artefacts arising from the inherent limitations of the technique (discussed in Section \ref{sec:Intro}). The second stems from the difficulty of manually annotating live FM data, especially when the FI of the probe fluctuates from frame to frame, thus introducing the additional challenge of retaining comprehensive temporal information from successive frames within a biologically meaningful 2D representation of the recording.

To address these issues, we develop a responsible, explainable and fully tractable methodology that deliberately avoids relying entirely on opaque DL models (as the ones mentioned in Section \ref{sec:DLapproach}). Instead, we focus on generating output data that are (1) straightforward to inspect, interpret, and annotate by biologists, and (2) optimally formatted to serve as input data for subsequent DL pipelines and downstream segmentation algorithms. These objectives converge towards the formulation of a generic framework that can be applied across diverse microscopy data scenarios. To illustrate our approach in this study, we have selected a unique dataset of FM recordings acquired from dynamic cardiac cell networks, which presents several challenges in terms of morphological complexity of the digitalised cellular structures, and the presence of noise and perturbations, thus being an excellent benchmark for this kind of applications.

\subsection{The Proposed Approach}\label{sec:framework}

\begin{figure}[ht!]
    \centering
    {\includegraphics[width=1.0\textwidth]{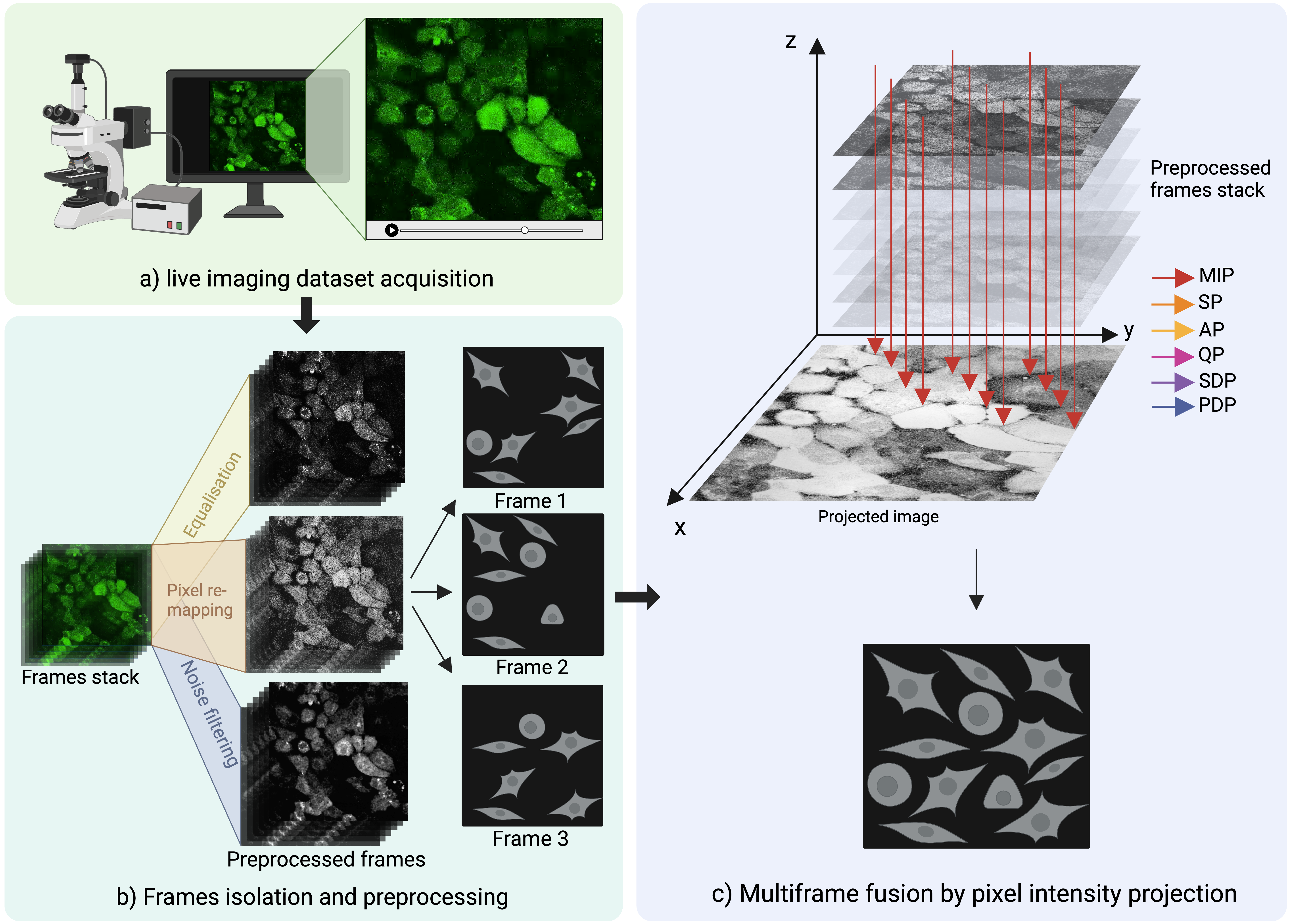}}
    \caption{Systematic pipeline for multi-temporal frames fusion in (a) live-sample FM datasets via combined (b) image processing and (c) z-projection methods.}
\label{fig:framework}
\end{figure}

Our proposed framework systematically combines image processing techniques (Section \ref{sec:preprocmethods}) with z-projection methods (Section \ref{sec:projection}) to perform multi-frame fusion in FM recordings. As graphically summarised in Figure \ref{fig:framework}, following data acquisition (Figure \ref{fig:framework}(a)), our approach isolates individual frames from video recordings to form multi-temporal stacks (Figure \ref{fig:framework}(b)).

The isolated frames are then individually preprocessed using a pipeline consisting of three sequential steps for quality enhancement: perform equalisation (see Section \ref{sec:equalisationMethods}), employ pixel intensity re-mapping (see Section \ref{sec:mappingMethods}), and apply noise filtering (see Section \ref{sec:filteringMethods}) (Figure \ref{fig:framework}(b)). Each preprocessed frame stack is subsequently fused using different projection techniques (described in Section \ref{sec:projection}), as illustrated in Figure \ref{fig:framework}(c). 

Our framework employs image enhancement and fusion operators not usually applied in the FM domain (refer to Section \ref{sec:projection}), guided by the innovative aim to use z-projections to reduce the dimensionality of temporally-resolved recordings of fluctuating fluorescent signals, similarly to how it is more commonly done for volumetric data \cite{zhou2025universal}. Nevertheless, multi-temporal frame fusion is fundamentally different from slicing 3D volumetric structures, as it poses the challenge to maintain temporal coherence across the frame evolution and within the fused image. To our knowledge, multi-temporal frames fusion has not yet been done for intermittently fluorescent data and has only been attempted preliminarily in our previous work \cite{caraffini2024towards,bib:costa2025AIiH}. However, in its current state, this technique should become an established practice to analyse molecular signalling experiments in FM.

The core focus of our investigation is to evaluate several configurations, generated by the combination of the three aforementioned preprocessing methods, followed by the appropriate projection technique. We do this by producing several variants of our framework, derived by cascading the processing and projection techniques reported in Section \ref{sec:methodlogy}, following the experimental procedure described in Section \ref{sec:expsetup}. We deliberately make our framework fully modular and easily extensible: additional techniques can be added to the pool of frames preprocessing and z-projection methods, to tailor combinations that effectively enhance and fuse images in diverse datasets. However, based on the specific case study discussed in this paper, we propose combinations of preprocessing and projection methods that worked best in our case (see Section \ref{sec:Recommendations}). We are confident that our approach will transfer well to many other datasets.

\section{Methodology}\label{sec:methodlogy}
The implementation of our framework is demonstrated through an extensive case study based on the following methodology.

\subsection{Live-Cell Imaging Dataset}\label{sec:dataset}

Data employed in this study consist of a live-cell imaging dataset comprising $91$ video recordings acquired from CLSM experiments on HL-1 CM networks, which have been widely characterised in previous studies  \cite{claycomb1998hl, caraffini2024towards, bib:costa2025AIiH, george2007alternative}. To visualise intracellular $Ca^{2+}$ signals, HL-1 networks were loaded with Fluo-3 AM, a fluorescent reporter used extensively in biological imaging of $Ca^{2+}$ signalling dynamics. Changes in Fluo-3 AM FI, associated with CM contraction, were recorded for $50s$ at $512\times512$ pixel resolution in AVI video format \cite{caraffini2024towards, bib:costa2025AIiH}. Each recording produced a multi-temporal frame stack of $150$ frames for subsequent processing.

\subsection{Frames Preprocessing Methods}\label{sec:preprocmethods}

\subsubsection{Equalisation}\label{sec:equalisationMethods}

We employ Contrast Limited Adaptive Histogram Equalisation (CLAHE) \cite{zuiderveld1994contrast}, an advanced HE technique designed to improve contrast in low-light images \cite{jin2025enhancing}. CLAHE enhances contrast locally, by equalising histograms of small non-overlapping tiles of size $t_s$, thus avoiding global noise amplification \cite{jin2025enhancing}. Calibrating $t_s$ is crucial to achieve successful contrast enhancement, preserve local details, and mitigate noise. Low $t_s$ values enhance fine details, but can amplify noise and introduce artefacts, whereas high $t_s$ values may over-smooth details and produce noise, similarly to global HE.

CLAHE uses a clip limit parameter $\alpha$ to cap the maximum bin height of each tile's histogram \cite{jin2025enhancing}, preventing excessive contrast amplification. Pixels exceeding this limit are redistributed uniformly within the tile, and tiles are then combined, typically via bilinear interpolation. Unless otherwise specified, $256$ gray levels are used \cite{zuiderveld1994contrast}. Higher $\alpha$ values produce stronger contrast enhancement, revealing fine details but increasing noise and artefacts, whereas lower $\alpha$ values yield milder contrast improvement with reduced risk of noise amplification but limited image enhancing effects \cite{zuiderveld1994contrast}.

CLAHE was developed to enhance contrast in medical images, which typically display low background-to-foreground light differences \cite{zuiderveld1994contrast}. It has become a widely used preprocessing method, often outperforming traditional HE in radiology \cite{nia2024medical}, as well as medical microscopy \cite{ningsih2020improving}, to support diagnosis. Therefore, we do not consider other traditional methods at this stage and instead use CLAHE in the two configurations defined in Table \ref{tab:methodsparams}.

\subsubsection{Pixel Intensity Re-Mapping }\label{sec:mappingMethods}
To enhance brightness and visibility, we apply GC (see Section \ref{sec:classicalpreproc}), an intensity adjustment technique that remaps pixel values via a non-linear operation (Equation \ref{eq:GC}) \cite{cao2018contrast}:

\begin{equation}\label{eq:GC}
    I_{t} = I_{m}\left(\frac{I_{i}}{I_{m}} \right)^{\gamma}
\end{equation}

where the new intensity $I_t$ is computed from the maximum intensity $I_m$ and the original pixel value $I_i$. The parameter $\gamma$ controls the final intensity: $\gamma < 1$ increases $I_t$, brightening; $\gamma > 1$ decreases $I_t$, darkening \cite{cao2018contrast}.

Pixel re-mapping adjusts image brightness to better match human perception, enhancing illumination and revealing details previously hidden in dark regions. By doing so, GC increases background-to-foreground contrast \cite{agarwal2024analysis}, making objects and fine details more distinguishable \cite{cao2018contrast}. Because it enhances contrast while preserving brightness, GC is widely used in medical imaging, especially to process CT and MRI scans before diagnosis \cite{agarwal2024analysis}. In this paper, we tested GC using two configurations, shown in Table \ref{tab:methodsparams}.

\subsubsection{Noise Filtering}\label{sec:filteringMethods}

Noise appears in images in many forms, from common salt-and-pepper patterns to more complex types that are harder to remove, such as Gaussian blur or gamma noise, the latter closely linked to laser-based acquisition methods like CLSM. Such noise can arise from the quality of the acquisition instrument or the specific processing methods applied \cite{bindal2022systematic}. In microscopy, multi-frame stacking methods, such as MIP, have been shown to produce motion artefacts in composite images generated from 3D intravital microscopy experiments \cite{lorenz2012digital}. These artefacts, which are caused by the misalignment of objects in subsequent frames due to movement, appear as smears around boundaries \cite{lorenz2012digital}, hindering their accurate recognition, which is crucial for segmentation. To reduce noise and artefacts in our images, we consider a series of noise filtering techniques, selecting methods widely tested on medical images. These are reported below. Parameter values employed for each method are instead reported in Table \ref{tab:methodsparams}.

\paragraph{Median blur (MB)} MB is a non-linear filter that replaces noisy pixels with the median intensity of neighbouring pixels in a region of size $\phi\_size$. It effectively removes salt-and-pepper noise while preserving edges and fine details, making it preferable than GF (see Section \ref{sec:classicalpreproc}).

\paragraph{Bilateral filtering (BF)} BF is a non-linear, local noise filter that replaces each noisy pixel with a weighted mean of intensities in a neighbourhood of diameter $d$ \cite{rao2023effective}. It removes noise while preserving edges, supporting segmentation, especially in CT and other medical images \cite{rao2023effective}. BF is controlled by $\sigma$ Colour ($\sigma_c$) and $\sigma$ Space ($\sigma_s$), which define the maximum colour and distance for averaging, with higher values yielding stronger smoothing.

\paragraph{Non-Local Means filtering (NF)} NF replaces noisy pixel values with a weighted average of other pixels, similar to BF, but uses non-local neighbourhoods across the image \cite{buades2005non}. NF is controlled by three parameters: $h$, which sets filter strength (higher values remove more noise but may remove fine details), template window size, and search window size, which define the regions used to compute weights. NF was chosen for its effectiveness in suppressing random noise in live-cell FM \cite{yang2010adaptive}.

\subsection{Projection Methods}\label{sec:projection}

We evaluate six computationally efficient and easy to implement z-projection techniques as simple yet effective tools to enhance image quality and perform multi-temporal frames fusion in FM.

\paragraph{Maximum Intensity Projection (MIP)} MIP is the most widely used image fusion technique \cite{chao2021new}. It compares the corresponding pixel intensities across frames in an image stack and assigns to each pixel the highest value (Figure \ref{fig:framework}(c)). This increases brightness and is particularly effective for low-light recordings \cite{singh2023review}, but ignores variability in the stack.
\paragraph{Average Projection (AP)} AP averages the intensities of the corresponding pixels in all frames \cite{mishra2015fusion}.
\paragraph{Sum Projection (SP)} SP sums the intensity values of corresponding pixels in all frames. The resulting images may have undesired pixel intensities that exceed the 0–255 range used for the 8-bit display (see Section \ref{sec:NR-IQAres}).
\paragraph{Peak-detection Projection (PDP)} PDP maps each pixel to the peak position where its maximum intensity occurs. Unlike MIP, PDP uses the $argmax$ function to return such a location. It is a shader algorithm classically used to determine, for example, object transparency or the most visually relevant depth along a ray in semi-transparent volume rendering \cite{3245305}.
\paragraph{Standard Deviation Projection (STDP)} STDP calculates the standard deviation of the intensity of the pixels along the z-dimension \cite{selinummi2009bright}, thus capturing the variability of the intensity across the entire stack. It has been used to project brightfield macrophage microscopy stacks, facilitating whole-cell segmentation and quantitative analysis \cite{selinummi2009bright}.

\paragraph{Quantile Projection (QP)} QP generates an image whose pixel values are the upper ($75^{th}$) percentile of the corresponding pixel intensities in all fused frames. In contrast to other dispersion measures such as the Inter-Quartile Range (IQR), this approach emphasises absolute intensity levels rather than variability. Methodologically, it is analogous to Median Projection methods (MDP), but with the quantile set to $0.75$ instead of $0.5$. To our knowledge, QP has not yet been applied to biomedical images, but we chose to employ it in this study as it preserves certain structural information of HL-1 networks compared to MDP and IQR, despite their broader use for medical imaging. We therefore recommend a more extensive use of QP in this domain. A visual comparison of the three mentioned methods is provided in Figure \ref{fig:compare_prjs}.

\subsection{Image Quality Evaluation Metrics}

In many practical scenarios, the unavailability of a GT reference image prevents the use of full-reference metrics, such as Peak Signal-to-Noise Ratio (PSNR) and Structural Similarity Index Measure (SSIM) for image quality assessment. To address the absence of GT images in our study, we employ No-Reference (NR) Image Quality Assessment (IQA) methods, which allow the extimation of perceptual quality from the image alone. This methods are especially useful in image compression, enhancement, and restoration \cite{dohmen2025similarity}. In medical imaging, NR-IQA methods have been evaluated for MRI \cite{dohmen2025similarity}, CT scans \cite{Gunawan2025}, and radiological images \cite{higashiyama2024investigation}, which often exhibit distortions, noise, and artefacts that obscure details and hinder diagnosis. Far fewer studies have examined NR-IQA to assess the quality of microscopy images.

NR-IQA methods use statistical features, Natural Scene Statistics (NSS), or learned models to detect noise, blur,  artefacts, and textures that substantially deviate from those typically observed in natural images, predicting quality scores that generally align with human perception \cite{dohmen2025similarity}. This study employs three standard NR-IQA metrics to assess the perceptual quality of FM images acquired from live HL-1 CM networks after processing (see Section \ref{sec:preprocmethods}) and fusion (see Section \ref{sec:projection}). The three metrics include:

\paragraph{Naturalness Image Quality Evaluator (NIQE)} This is a completely blind, model-based metric that does not require training using subjective quality scores \cite{mittal2012making}. It measures quality by quantifying the deviation of an image from  statistical regularities modelled in natural scenes, making it effective at identifying global distortions \cite{mittal2012making}. Because it focuses on image `naturalness', lower NIQE scores indicate higher quality, closer to natural images.

\paragraph{The Perception-based Image Quality Evaluator (PIQE)} This is a block-wise, no-reference metric that assesses perceptual quality by analysing local distortions \cite{venkatanath2015blind}. It divides the image into non-overlapping blocks, computes spatial activity and deformations, and aggregates these into a score, with lower scores indicating higher quality. This makes PIQE sensitive to localised degradations such as noise and blur, while remaining computationally efficient \cite{venkatanath2015blind}. 

\paragraph{The Blind/Referenceless Image Spatial Quality Evaluator (BRISQUE)} This is a learning-based NR-IQA method that models NSS in the spatial domain \cite{mittal@BRISQUE}. It extracts features, capturing deviations from expected NSS and uses a trained regressor to predict perceptual quality. BRISQUE effectively captures local and global distortions, where lower scores indicate higher quality. Unlike NIQE, it requires a model trained on images with subjective quality scores, which enables the accurate prediction of human-perceived quality across diverse real-world datasets.

\vspace{1em}
Before evaluation, images were min–max normalised to [0, 255] for conversion into 8-bit format to ensure metric compatibility and reduce conversion-related information loss, especially for images projected using SP, where intensity values can exceed the 8-bit range. Best images and best preprocessing-projection methods combinations identified by NIQE, PIQE and BRISQUE were isolated and compared with an expert experimentalist researcher's perception, to assess how well each metric perceives microscopy image quality. As these metrics were developed by studying NSS, their suitability for microscopy data, whose properties differ markedly, remains uncertain and is evaluated and further discussed in Sections \ref{sec:NR-IQAres} and \ref{Sec:Discussion}.

\section{Experimental setup}\label{sec:expsetup}

Using our general framework (Section \ref{sec:framework}) in multiple configurations of operators described in Section \ref{sec:methodlogy} to process the dataset described in Section \ref{sec:dataset}, we generate several scenarios for evaluation (Algorithm \ref{alg:framework}).

\begin{algorithm}
\scriptsize
\caption{Pipelines generation and execution}
\label{alg:framework}
\begin{algorithmic}[0] 
\State Fetch $\mathcal{V} = \{v^{(1)}, \dots, v^{(n)}\}$  \Comment{The set of $n=91$ AVI videos fr this study (Section \ref{sec:dataset})}
\State Fetch operators' parameters \Comment{Table \ref{tab:methodsparams}}
    \State $\mathcal{E} = \{\text{CL} , \text{CH}\}$\Comment{Initialise the equalisation operator set (Section \ref{sec:equalisationMethods})}
    \State $\mathcal{G} = \{\text{GL} , \text{GH}\}$\Comment{Initialise the pixel mapping operators set (Section \ref{sec:mappingMethods})}
    \State $\mathcal{F} = \{\text{MF}, \text{BF}, \text{NF}\} $\Comment{Initialise the filter operators set (Section \ref{sec:filteringMethods})}
    \State $\mathcal{P} = (\text{SP}, \text{AP}, \text{MIP}, \text{PDP}, \text{SDP}, \text{QP})$\Comment{Initialise the projection operators set (Section \ref{sec:projection})}
 
    \State $\mathcal{S} \gets\text{Combine}\left(\mathcal{E},\mathcal{G},\mathcal{F}\right)$ \Comment{Generate preprocessing sequences; the code is available at \cite{bib:ProjectionZenodoRepo}}
\State $\tilde{P}\gets\emptyset$\Comment{To store normalised projected 2D outcomes}
\For{each $v \in \mathcal{V}$}
    \For{each $s \in \mathcal{S}$}
    \State  $v^s\gets$ apply $s$ frame-wise to $v$ 
    \For{each $p \in \mathcal{P}$}
        \State $\tilde{\mathcal{P}} \gets$ apply $p$ to $v^s$ and store after normalising its output
    \EndFor
\EndFor
\EndFor
\State Return $\tilde{P}$.
\end{algorithmic}
\end{algorithm}

Note that a preprocessing sequence in $\mathcal{S}$ can be: a single operator taken from one of the sets $\mathcal{E}$, $\mathcal{G}$, or $\mathcal{F}$; a pair of operators, where each operator comes from a different one of these sets (in any order); a triple of operators, again with each operator taken from a different set among $\mathcal{E}$, $\mathcal{G}$, and $\mathcal{F}$ (in any order). This generates 111 preprocessing configurations, which we assess in this study in combination with each of the six projection operators in $\mathcal{P}$ for a total of 666 pipelines.

For the sake of reproducibility, we report the parameter setting of the operators employed in Table \ref{tab:methodsparams} and we make the code to replicate this experimental setup available in \cite{bib:ProjectionZenodoRepo}. The code can be easily extended to run additional pipelines or modified to select only a desired one.

\begin{table}[ht!]
        \centering
        \scriptsize
        \begin{adjustbox}{max width = \textwidth}
        \begin{tabular}{lcl}
        \hline\hline
        \textbf{Method} & \textbf{Acronym} & \textbf{Parameters} \\
        \hline
        \multirow{2}{*}{Equalisation} 
        & CL 
        & $\alpha = 1$, $t_s = (16, 16)$ \\
        & CH
        & $\alpha = 4$, $t_s = (4, 4)$ \\
        \hline
        \multirow{2}{*}{Pixel re-mapping} 
        & GL 
        & $\gamma = 0.75$ \\
        & GH
        & $\gamma = 1.25$ \\
        \hline
        \multirow{3}{*}{Noise filtering} 
        & MB
        & $\phi\_size = 5$ \\
        & BF
        & $d = 3$, $\sigma_c = 25$, $\sigma_s = 50$ \\
        & NF
        & $h = 10$, TWS $= 3$, SWS $= 7$ \\
        \hline\hline
        \end{tabular}
        \end{adjustbox}
        \caption{Acronyms and parameter values for each processing method used in the study.}
        \label{tab:methodsparams}
    \end{table}

Processing our $n = 91$ recordings with all pipelines in Algorithm \ref{alg:framework} yields 60,606 results for analysis. This extensive setup facilitates a thorough examination of various scenarios, allowing for meaningful conclusions.

\section{Results}\label{sec:results}

\subsection{No-reference Image Quality Assessment}{\label{sec:NR-IQAres}}

Table \ref{tab:Descr_stats_scores} reports descriptive statistics for PIQE, NIQE and BRISQUE scores computed on the fully preprocessed and projected HL-1 dataset ($n = 60,606$). Overall, PIQE and NIQE scores were shown to largely fall within the 0 to 100 range, with lower values indicating higher perceptual image quality and 100 indicating perfect quality. BRISQUE scores exhibited a broader numerical range compared to PIQE and NIQE, spanning from extremely high values, up to 181.23 for the worst image, to extremely low values, down to $-13.86$ for the best image reported in the study (Table \ref{tab:Descr_stats_scores}). 

\begin{table}[ht]
        \centering
        \begin{tabular}{c|c|c|c}
        \hline
            Descriptive Stats  & PIQE & NIQE & BRISQUE \\
            \hline
            Min -- Max & 0.59 -- 79.76 &  8.03 -- 50.42 &  -13.86 -- 181.23 \\      
            Mean $\pm$ SD & $21.83\pm18.78 $  & $18.25\pm6.78$  & $40.61\pm36.83$ \\ 
            Median (IQR) & 16.46 (29.45) & 16.29 (6.03) & 31.28 (47.71)\\
        \hline
        \end{tabular}
        \caption{Descriptive statistics for PIQE, NIQE and BRISQUE scores on the projected HL-1 dataset. Results are reported collectively, without separating data by preprocessing–projection method combinations.}
        \label{tab:Descr_stats_scores}
    \end{table}

\begin{figure}[ht]
    \centering
    {\includegraphics[width=1.0\textwidth]{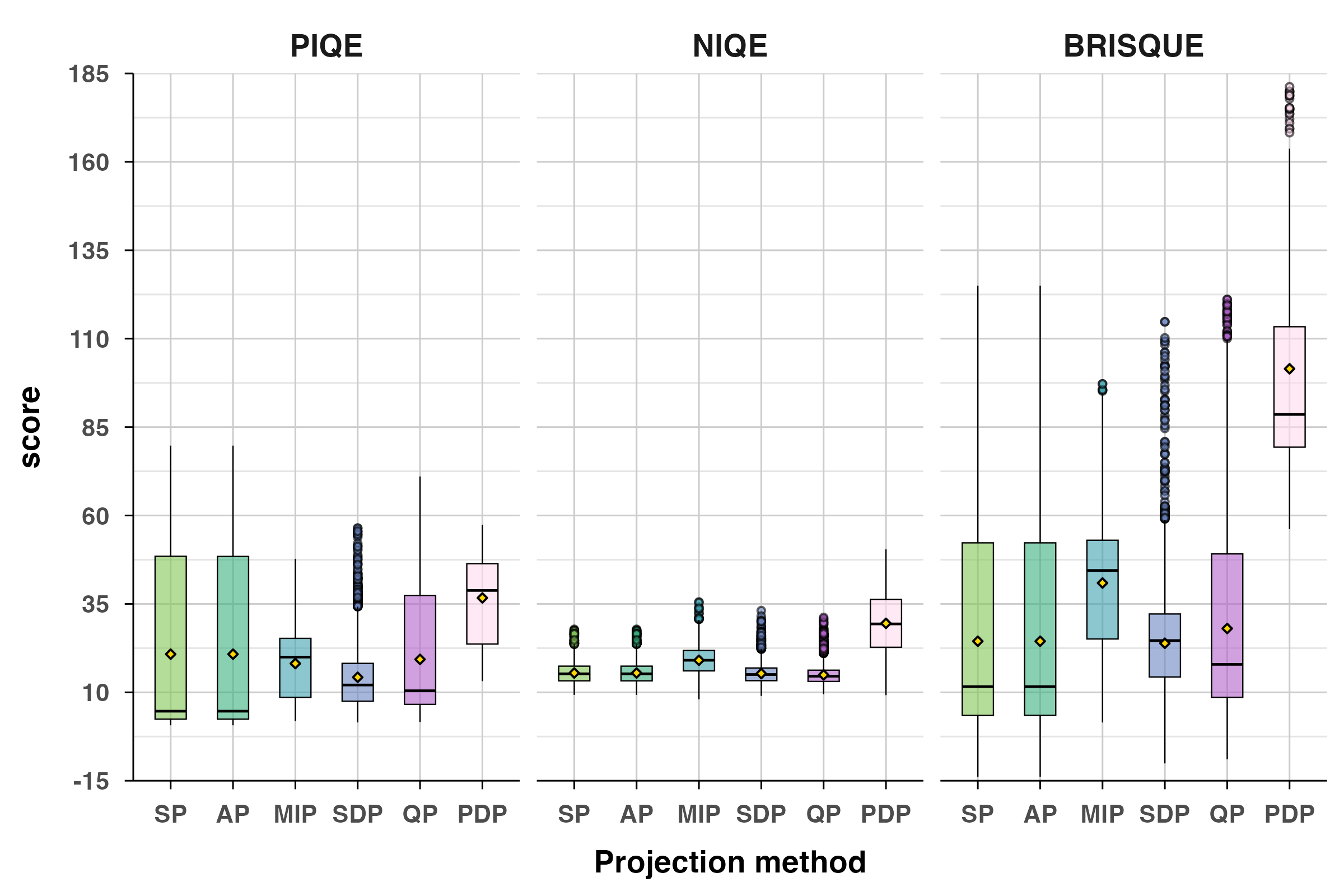}}
    \caption{Box plot of NR-IQR scores by projection method. Yellow-filled diamond-shaped dots within the boxes indicate mean values for each projection group.}
\label{fig:stats_box}
\end{figure}

\begin{figure}[ht!]
\begin{subfigure}{0.32\textwidth}
    \centering
    \scriptsize
    {P = 0.59} \hspace{2.5em} {P = 0.78} \\ [0.3em]
    \includegraphics[width=0.48\linewidth]{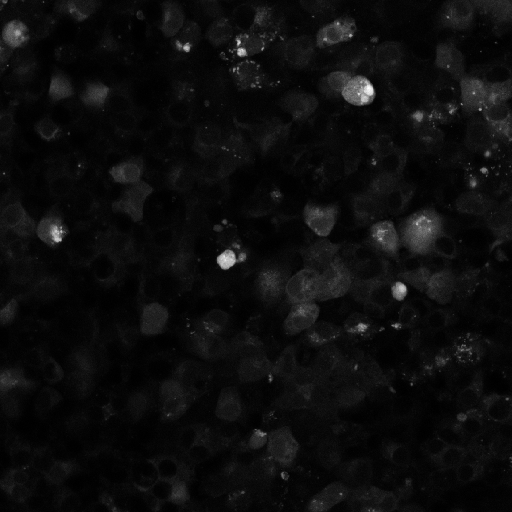}
    \includegraphics[width=0.48\linewidth]{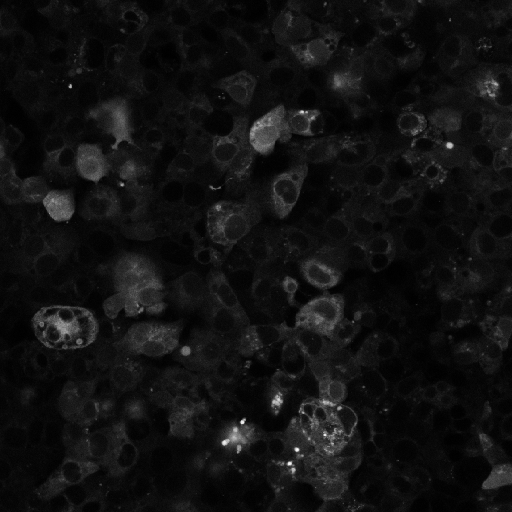}
\end{subfigure}
\begin{subfigure}{0.32\textwidth}
    \centering
    \scriptsize
    {P = 16.46} \hspace{2.5em} {P = 16.46} \\[0.3em]
    \includegraphics[width=0.48\linewidth]{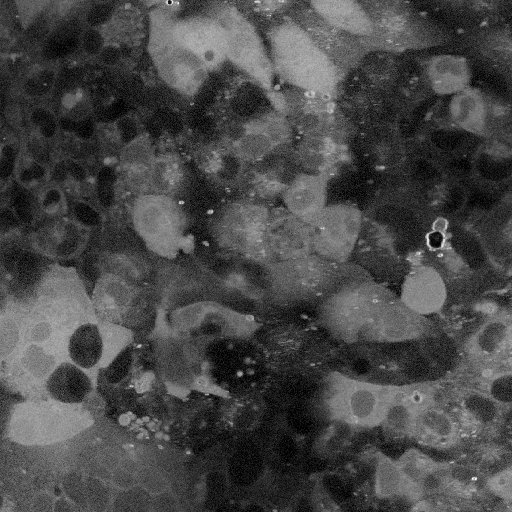}
    \includegraphics[width=0.48\linewidth]{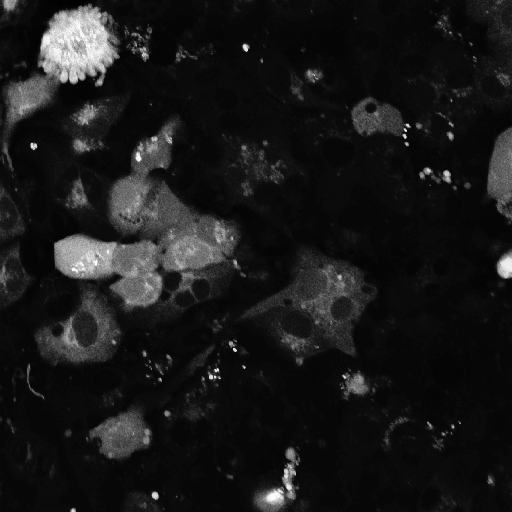}
\end{subfigure}
\begin{subfigure}{0.32\textwidth}
    \centering
    \scriptsize
    {P = 79.07} \hspace{2.5em} {P = 79.76} \\[0.3em]
    \includegraphics[width=0.48\linewidth]{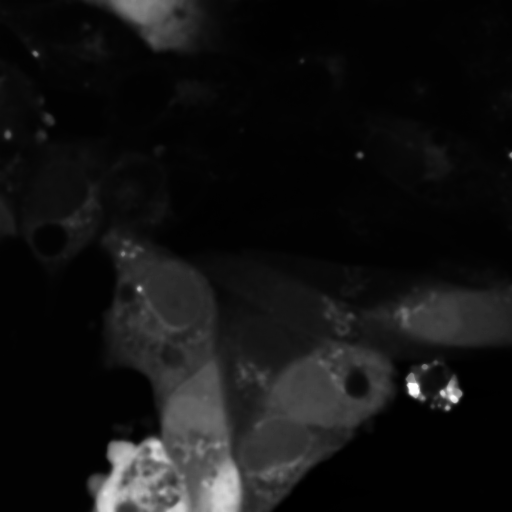}
    \includegraphics[width=0.48\linewidth]{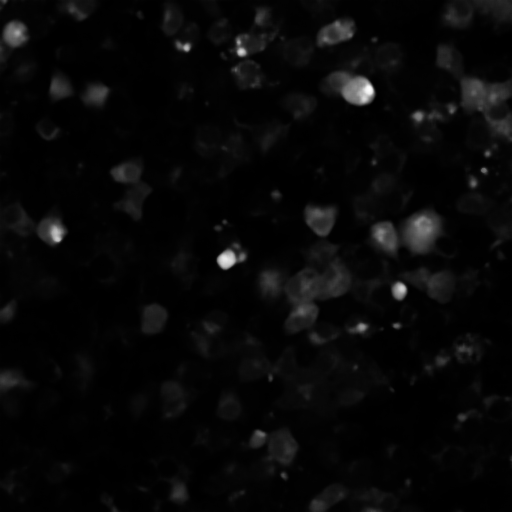}
\end{subfigure} 

\vspace{0.3em}
\begin{subfigure}{0.32\textwidth}
    \centering
    \scriptsize
    {N = 8.03} \hspace{2.5em} {N = 8.51} \\[0.3em]
    \includegraphics[width=0.48\linewidth]{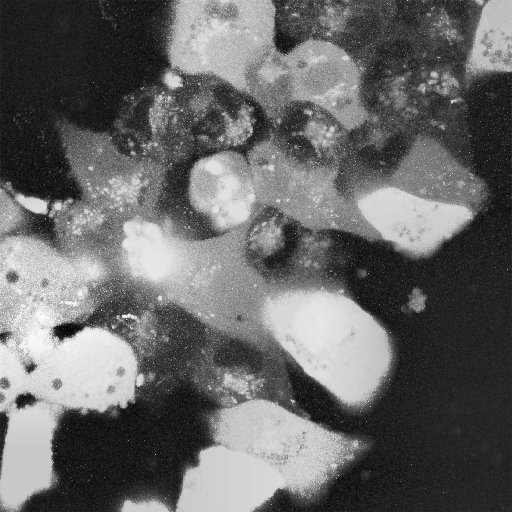}
    \includegraphics[width=0.48\linewidth]{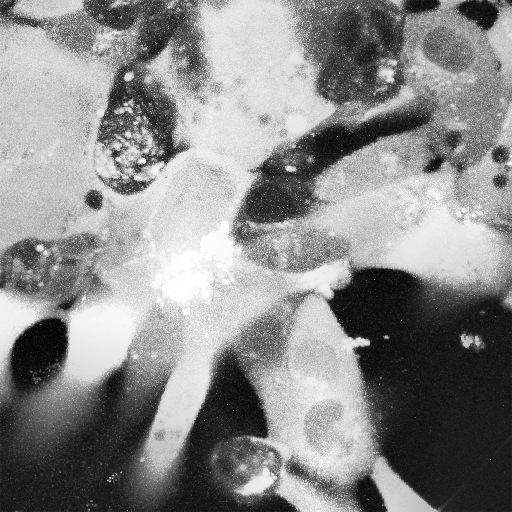}
\end{subfigure}
\begin{subfigure}{0.32\textwidth}
    \centering
    \scriptsize
    {N = 16.29} \hspace{2.5em} {N = 16.29} \\[0.3em]
    \includegraphics[width=0.48\linewidth]{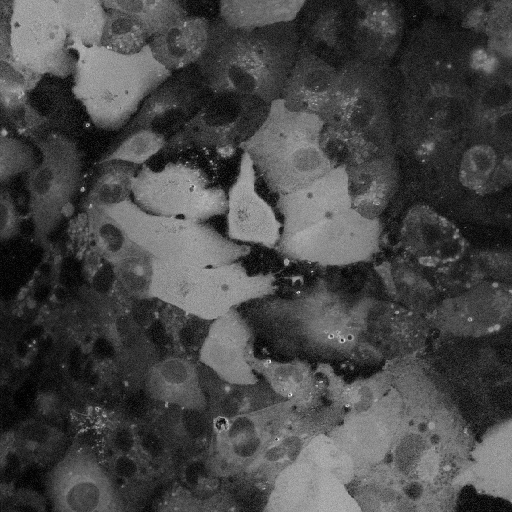}
    \includegraphics[width=0.48\linewidth]{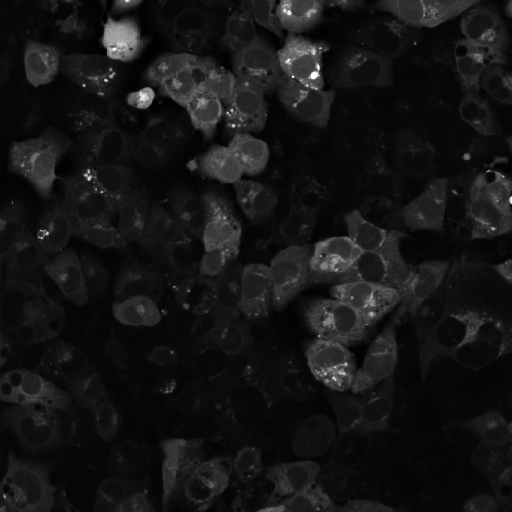}
\end{subfigure}
\begin{subfigure}{0.32\textwidth}
    \centering
    \scriptsize
    {N = 48.83} \hspace{2.5em} {N = 50.42} \\[0.3em]
    \includegraphics[width=0.48\linewidth]{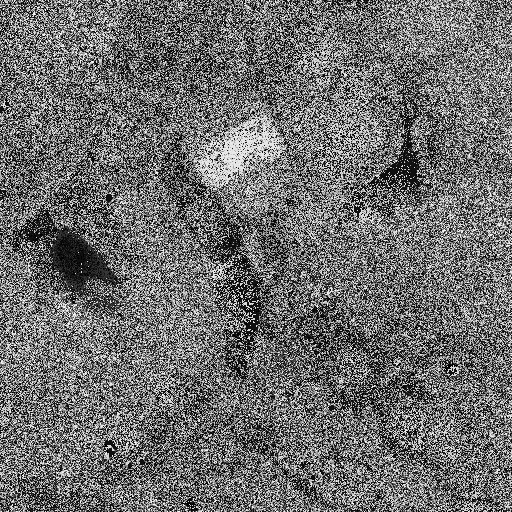}
    \includegraphics[width=0.48\linewidth]{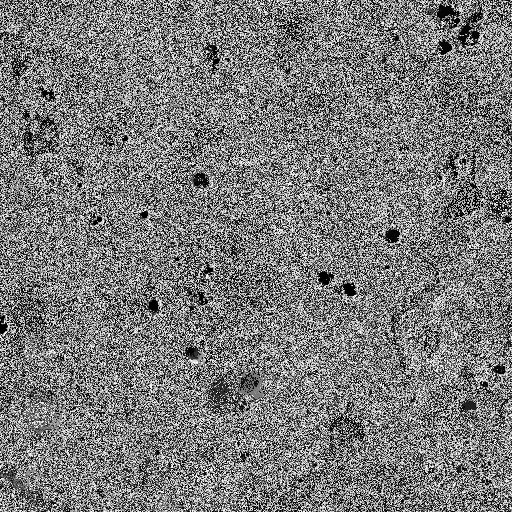}
\end{subfigure}

\vspace{0.3em}
\begin{subfigure}{0.32\textwidth}
    \centering
    \scriptsize
    {B = -13.86} \hspace{2.0em} {B = -10.14} \\[0.3em]
    \includegraphics[width=0.48\linewidth]{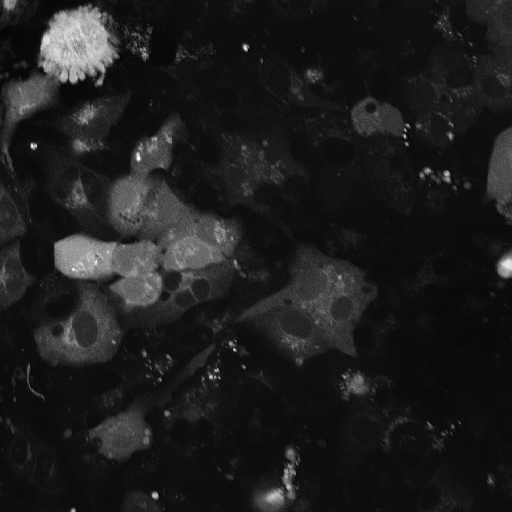}
    \includegraphics[width=0.48\linewidth]{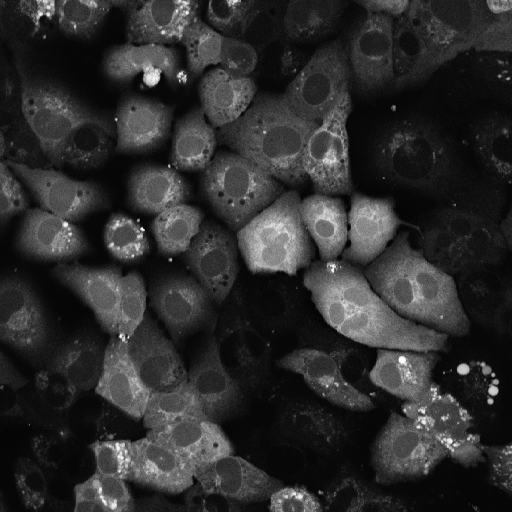}
    \caption{High Quality}
\end{subfigure}
\begin{subfigure}{0.32\textwidth}
    \centering
    \scriptsize
    {B = 31.28} \hspace{2.0em} {B = 31.28} \\[0.3em]
    \includegraphics[width=0.48\linewidth]{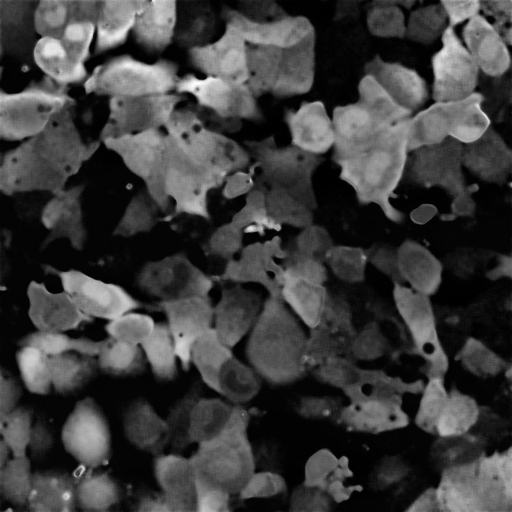}
    \includegraphics[width=0.48\linewidth]{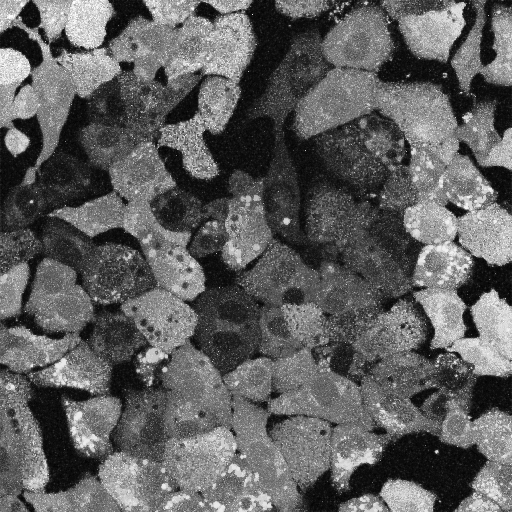}
    \caption{Median Quality} 
\end{subfigure}
\begin{subfigure}{0.32\textwidth}
    \centering
    \scriptsize
    {B = 179.92} \hspace{2.0em} {B = 181.23} \\[0.3em]
    \includegraphics[width=0.48\linewidth]{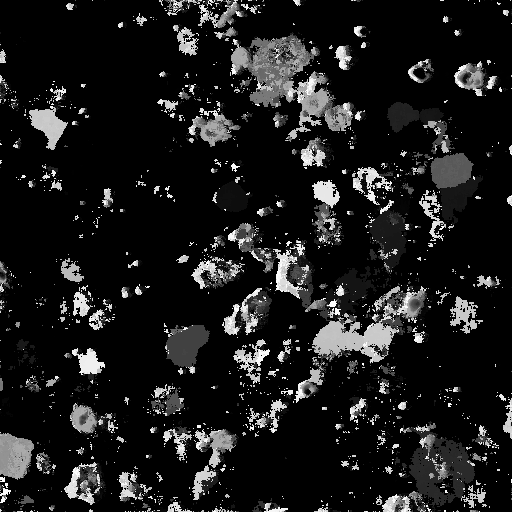}
    \includegraphics[width=0.48\linewidth]{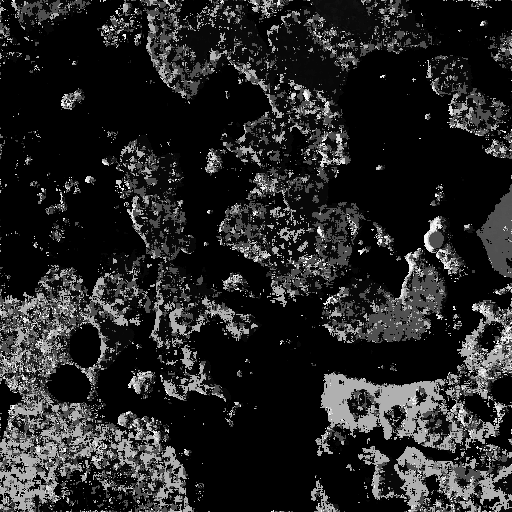}
    \caption{Low Quality}
\end{subfigure}
    
\caption{Qualitative analysis of NR-IQA metrics image quality evaluation performance on the HL-1 dataset. Individual rows show examples of images scoring a) the lowest, b) median and c) the highest PIQE (P), NIQE (N) and BRISQUE (B) scores across the full dataset, corresponding to a) high, b) medium and c) low scores assigned by each metric.}
\label{fig:scores_visual}
\end{figure}

Stratifying the data by the six projection methods (Figure \ref{fig:stats_box}) linked abnormally high BRISQUE values to images processed with PDP. These images also showed generally higher PIQE and NIQE scores than the other techniques. Qualitative inspection of images with the worst BRISQUE scores confirmed corruption (Figure \ref{fig:scores_visual}(c)). As a result, PDP images were unusable for GT annotation and were discarded from further analysis.

Given the large number of outliers in the box-plot analysis shown in Figure \ref{fig:stats_box}, we prioritized the median over the mean to select an optimal projection method for frames fusion and image quality enhancement on the HL-1 dataset. Overall, SP and AP performed best on PIQE and BRISQUE, yielding the lowest median values (PIQE$_{SP}$, PIQE$_{AP}$ = 5.03; BRISQUE$_{SP}$, BRISQUE$_{AP}$ = 12.08), with the minimum BRISQUE score (-13.86) found in the AP group (Figure \ref{fig:stats_box}). SP and AP images showed similar performance patterns (mean, median, and distributions) across all three NR-IQA metrics (Figure \ref{fig:stats_box}), indicating comparable behaviour on the HL-1 dataset. However, independent paired t-tests and effect-size analyses across PIQE, NIQE, and BRISQUE (available in \cite{bib:ProjectionZenodoRepo}) revealed several statistically significant differences and high effect sizes between SP and AP for some preprocessing-projection method combinations. Thus, while similar, these methods are not fully interchangeable and should both be considered for downstream analysis.

QP, directly followed by SDP, were associated with the lowest median NIQE scores (i.e., NIQE$_{QP}$ = 14.55, NIQE$_{SDP}$ = 15.01), despite demonstrating a lower performance than SP and AP methods based on PIQE and BRISQUE scores. Finally, images processed using MIP were associated with lower quality scores across all metrics compared to SP, AP, QP and SDP.

To complement the quantitative analysis of NR-IQA scores and assess whether the chosen image quality metrics effectively correlate with changes in the visual quality of our microscopy images, we conducted a complementary qualitative analysis on an exemplary subset of the HL-1 dataset (N = 18). PIQE, NIQE and BRISQUE were individually sorted in ascending order to identify exemplary images scoring the closest values to the minima, median and maxima identified in Table \ref{tab:Descr_stats_scores}. Results of this analysis are summarised in Figure \ref{fig:scores_visual}. PIQE assigned the highest values (i.e., P = 79.07 and P = 79.76) to images characterised by high levels of blur (Figure \ref{fig:scores_visual}c), penalising decreases in image sharpness and losses of fine details over other signs of image quality deterioration, including the high salt and pepper noise and low brightness observed in high and median quality images for PIQE (Figure \ref{fig:scores_visual}b). Differently, NIQE and BRISQUE were shown to penalise images that were substantially degraded by the specific preprocessing pipeline applied, primarily images belonging to the PDP group, associated with the highest values for all three metrics (Figure \ref{fig:stats_box}). Both NIQE and BRISQE were proven sensitive to alterations of the original images, leading to the generation of artefacts introduced by the preprocessing pipeline and previously not observed within the raw frames. Examples of these alterations are visible from the exemplary median quality images reported for the two metrics (Figure \ref{fig:scores_visual}b). NIQE was also shown to penalise low image contrast, considering images characterised by high foreground-to-background contrast of higher quality than images characterised by low brightness and low illumination variability between cellular and background structures.

\begin{figure}[ht!]
\begin{subfigure}{0.24\textwidth}
    \centering
    \scriptsize
    {P = 0.59} \\ [0.2em]
    \includegraphics[width=\linewidth]{Images/PIQE_mean_117B_F_CL_NF_GL.png} \\ [0.2em]
    {\textbf{AP\_CL\_NF\_GL}} 
\end{subfigure}
\begin{subfigure}{0.24\textwidth}
    \centering
    \scriptsize
    {P = 9.60} \\ [0.2em]
    \includegraphics[width=\linewidth]{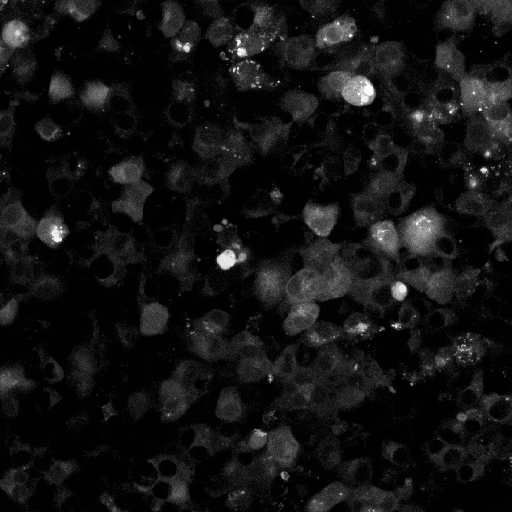} \\[0.2em] 
    {\textbf{QP\_CL\_NF}} 
\end{subfigure}
\begin{subfigure}{0.24\textwidth}
    \centering
    \scriptsize
    {P = 14.33} \\ [0.2em]
    \includegraphics[width=\linewidth]{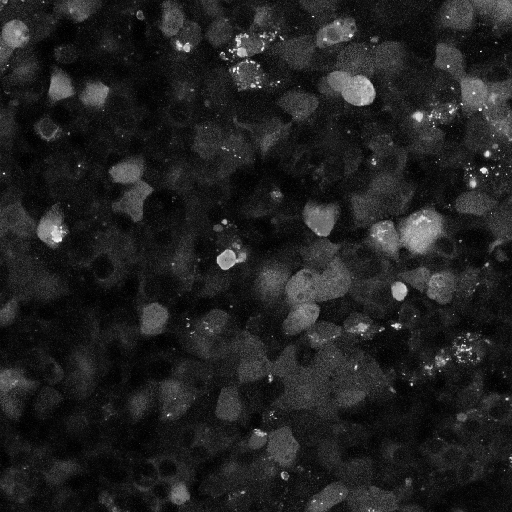} \\[0.2em] 
    {\textbf{SDP\_GL\_CL\_BF}} 
\end{subfigure}
\begin{subfigure}{0.24\textwidth}
    \centering
    \scriptsize
    {P = 14.29} \\ [0.2em]
    \includegraphics[width=\linewidth]{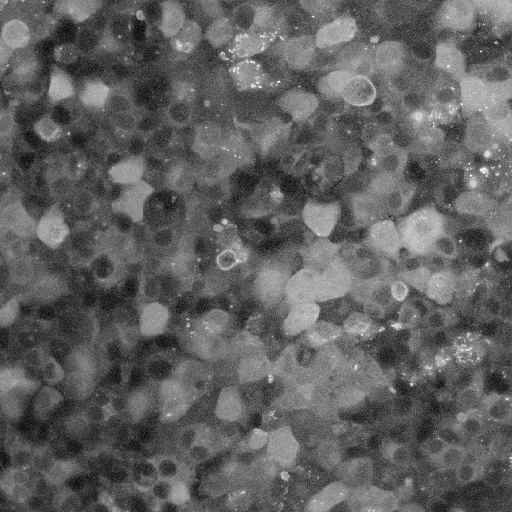} \\[0.2em] 
    {\textbf{SDP\_GH\_CH}} 
\end{subfigure}
\vspace{0.5em}
\begin{subfigure}{0.24\textwidth}
    \centering
    \scriptsize
    {N = 8.03} \\ [0.2em]
    \includegraphics[width=\linewidth]{Images/NIQE_max_194_F_BF_CH_GH.png} \\ [0.2em]
    {\textbf{MIP\_BF\_CH\_GH}} 
\end{subfigure}
\begin{subfigure}{0.24\textwidth}
    \centering
    \scriptsize
    {N = 12.41} \\ [0.2em]
    \includegraphics[width=\linewidth]{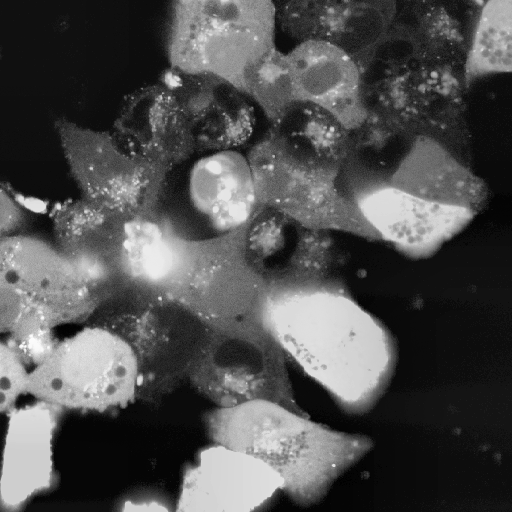} \\[0.2em] 
    {\textbf{QP\_CH\_GH\_NF}} 
\end{subfigure}
\begin{subfigure}{0.24\textwidth}
    \centering
    \scriptsize
    {N = 8.73} \\ [0.2em]
    \includegraphics[width=\linewidth]{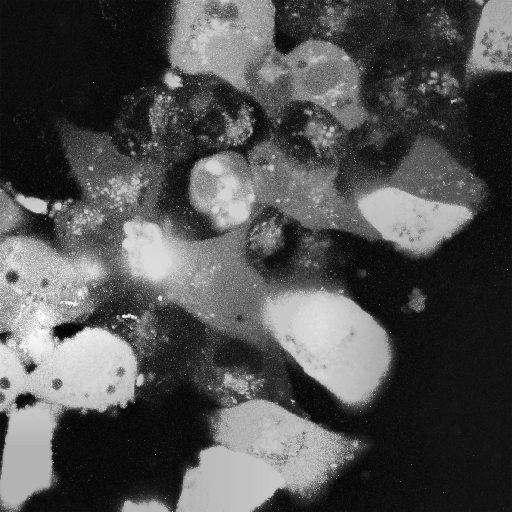} \\[0.2em] 
    {\textbf{MIP\_BF\_CH\_GL}} 
\end{subfigure}
\begin{subfigure}{0.24\textwidth}
    \centering
    \scriptsize
    {N = 8.50} \\ [0.2em]
    \includegraphics[width=\linewidth]{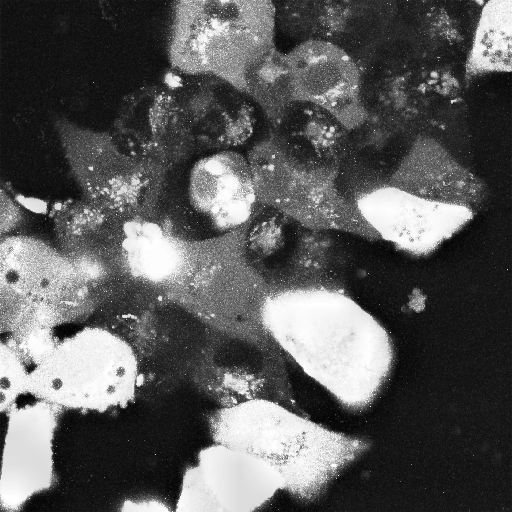} \\[0.2em] 
    {\textbf{MIP\_BF\_GH\_CL}} 
\end{subfigure}
\vspace{0.5em}
\begin{minipage}{0.24\textwidth}
    \centering
    \scriptsize
    {B = -13.86} \\ [0.2em]
    \includegraphics[width=\linewidth]{Images/BRISQUE_mean_57B_F_NF_GH_CH.png} \\ [0.2em]
    {\textbf{AP\_NF\_GH\_GH}}
    \caption*{a) HPQ - Scores}
\end{minipage}
\hspace{-0.6em}
\begin{minipage}{0.75\textwidth}
    \centering
    \scriptsize
    {B = 33.37} \hspace{6.5em} {B = 57.69} \hspace{6.5em}
    {B = 55.27} \\ [0.2em]
    \includegraphics[width=0.32\linewidth]{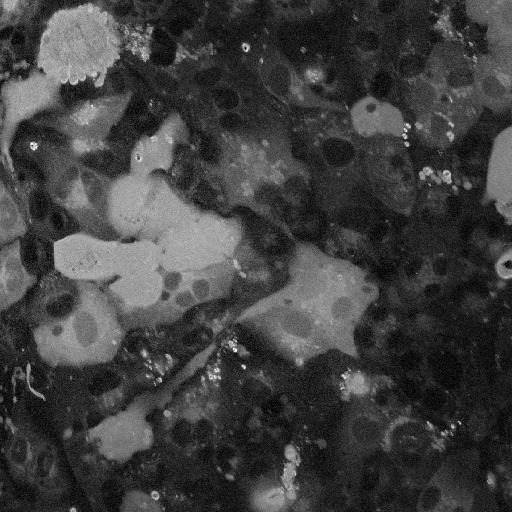} 
    \includegraphics[width=0.32\linewidth]{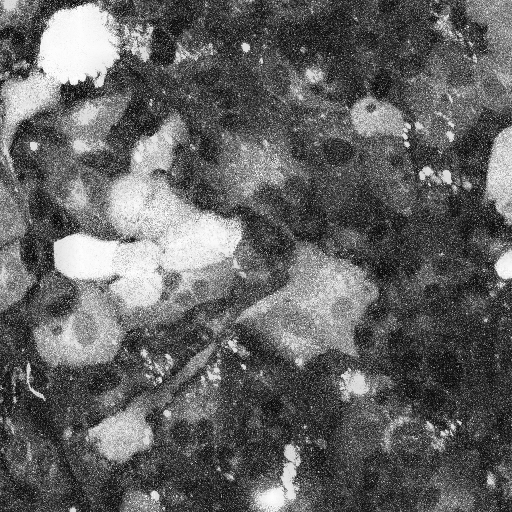} 
    \includegraphics[width=0.32\linewidth]{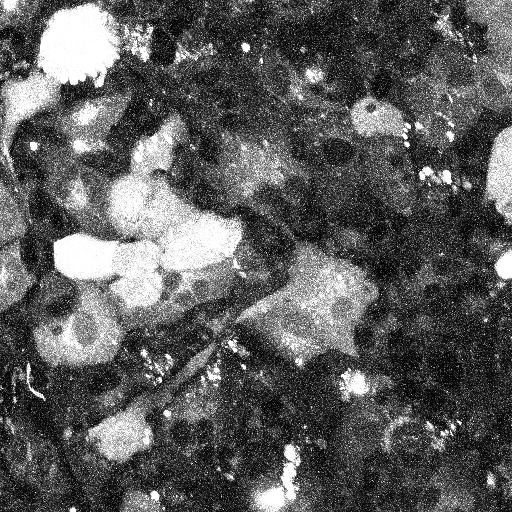} \\ [0.2em]
    {\textbf{SDP\_GH\_CH}} \hspace{1.5em} {\textbf{MIP\_BF\_GH\_CL}} \hspace{3em}
    {\textbf{MIP\_BF\_GH}} \\ [0.2em]
    \caption*{b) HPQ - Experimentalist}
\end{minipage}

\caption{Comparison of HPQ as assessed by a) NR-IQA scores and by b) an expert experimentalist. Panel b shows three exemplary images of preprocessing-projection methods yielding comparatively better results than the best image assigned through the calculation of PIQE (P), NIQE (N), and BRISQUE (B) scores. Exemplary images are ordered, from slightly better to significantly better than the P, N and Q maxima according to the experimentalist's perception of the overall image quality and usefulness for GT annotation.}
\label{fig:Scores_vs_bio}
\end{figure}

Highest quality images identified by the lowest PIQE, NIQE and BRISQUE scores were subjected to the expertise of an experimentalist researcher with extensive experience of working with $Ca^{2+}$ signals from HL-1 networks, which empirically determined the existence of incongruencies between the perceived quality assigned by each NR-IQA score and the visual quality and usefulness of the image for GT annotation (see Section \ref{sec:GTcomp}). Figure \ref{fig:Scores_vs_bio} reports the results of this analysis, comparing images identified of High Perceived Quality (HPQ) by PIQE, NIQE and BRISQUE scores (Figure \ref{fig:Scores_vs_bio}a) with images considered to have better perceived quality by the expert. It is worth noticing that, while HPQ images identified by the experimentalist substantially outperform the quality of the best HPQ images identified by PIQE and BRISQUE, we found agreement between the HPQ images identified by NIQE and the ones chosen by the experimentalist, with three out of four images scoring similar NIQE scores to the HPQ images selected by the experimentalist. Oppositely, PIQE and BRISQUE showed incongruencies between the score assigned to the best (Figure \ref{fig:scores_visual}a) and median quality images (Figure \ref{fig:scores_visual}b) and the translation of these scores into the effective perceived quality of the images. Therefore, both scores assigned HPQ values to low brigthness images characterised by poor foreground-to-background contrast and extremely impaired visibility of regions of cellular density and most importantly of cellular boundaries, essential to allow the annotations of HL-1 images.

\begin{figure}[ht!]
    \centering
    {\includegraphics[width=1.0\textwidth]{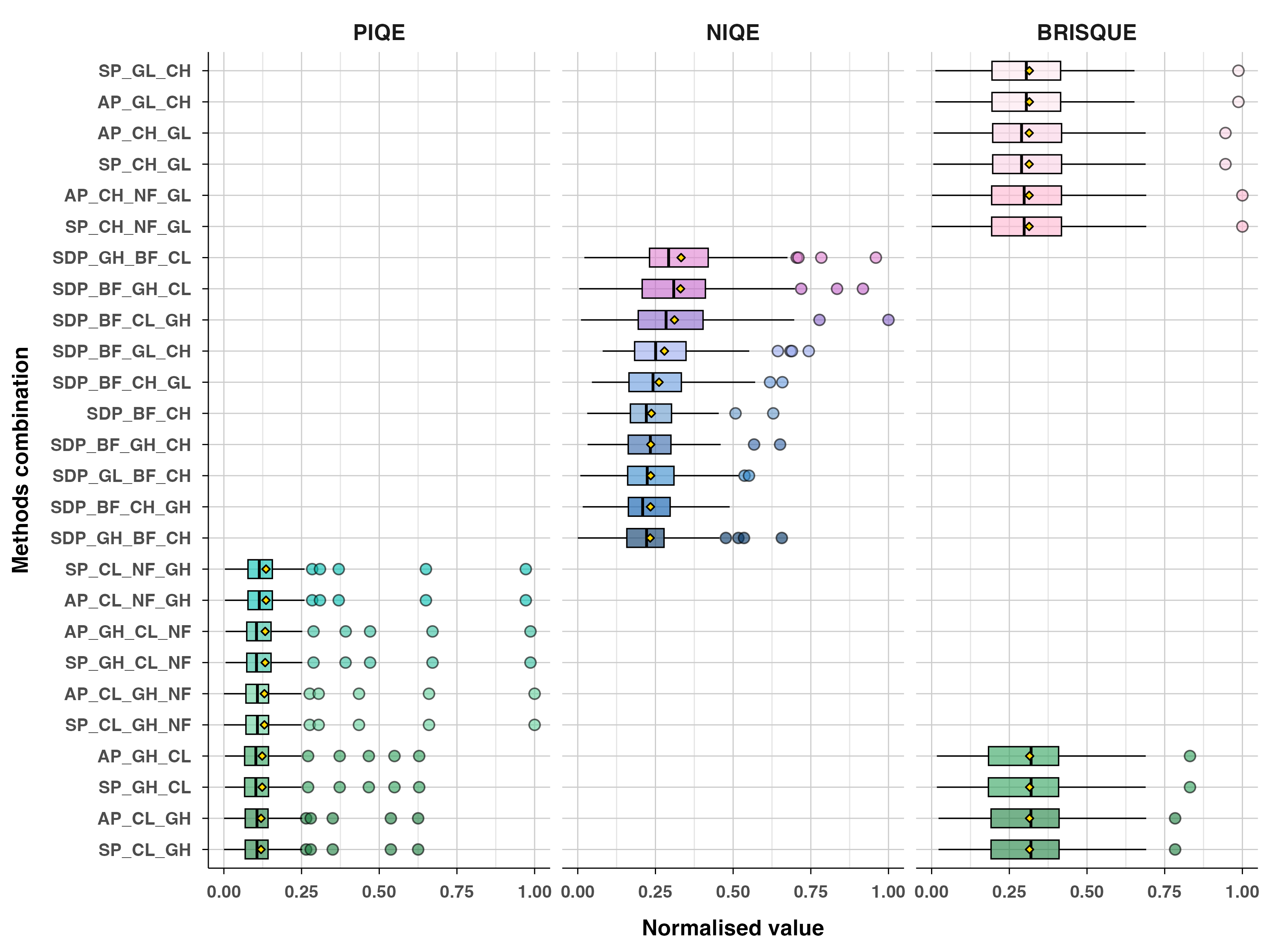}}
    \caption{Box-plot highlighting the 10 best preprocessing-projection algorithms combinations identified by PIQE, NIQE and BRISQUE scores. Data were normalised to allow a fair comparison between the three metrics. As before, yellow diamond-shaped dots within the boxes indicate mean values across the 77 images within the dataset, averaged across the specified preprocessing-projection method applied.}
\label{fig:best_averaged}
\end{figure}

Finally, Figure \ref{fig:best_averaged} shows the 10 best combinations of preprocessing-projection methods calculated as an aggregated average of the whole HL-1 dataset for each NR-IQA metric. The graph shows concordance between the best four methods identified by PIQE and BRISQUE, but no concordance of the two metrics with NIQE, with BRISQUE mainly assigning HPQ scores to images processed using SP and AP methods, and NIQE assigning HPQ values to images projected using SDP methods, in better accordance with the methods valued of HPQ by the experimentalist (see Figure \ref{fig:Scores_vs_bio} and Section \ref{sec:GTcomp}).

\subsection{GT comparison}\label{sec:GTcomp}

Samples of images generated using our pipeline were shown to an expert experimentalist for the selection of preprocessing-projection method combinations which could provide comprehensive, useful and different information about the cardiac cell networks imaged for GT annotation. Selected methods are shown in Figure \ref{fig:selected_prjs}. GT annotations generated using the selected projections were compared with annotations generated from composite grayscale images obtained using conventional frames stacking and averaging methods \cite{caraffini2024towards}. Due to visualisation constraints in old composite images due to low brightness and impaired image quality, only 77 images were analysed at this stage. New GT ($GT_{n}$) and old GT ($GT_{0}$) annotations were compared using two levels of analysis:

\begin{figure}[ht!]
    \centering

    \begin{subfigure}[b]{0.24\textwidth}
        \centering
        \includegraphics[width=\linewidth]{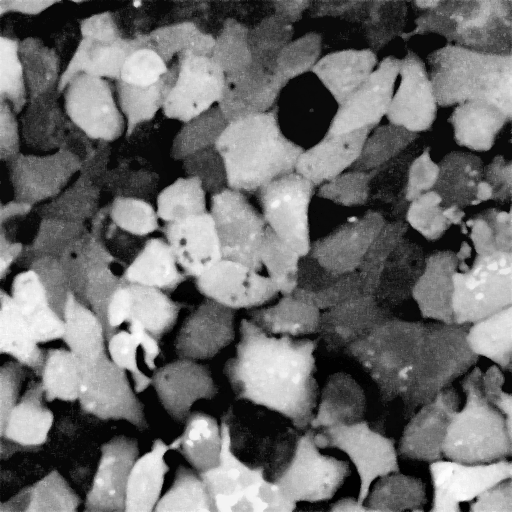}
        \caption{\scriptsize MIP\_CH\_GH\_MB}
        \label{fig:a}
    \end{subfigure}%
    \hfill
    \begin{subfigure}[b]{0.24\textwidth}
        \centering
        \includegraphics[width=\linewidth]{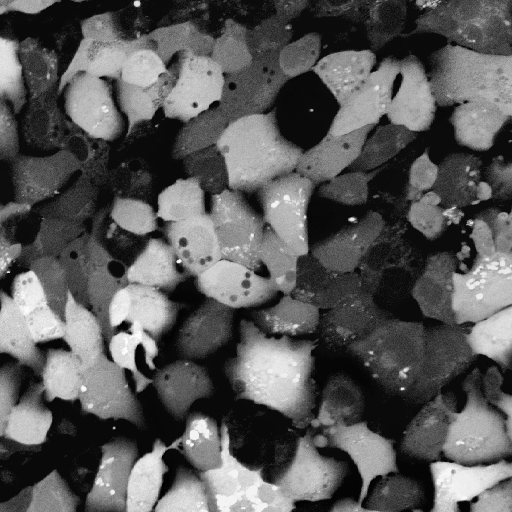}   
        \caption{\scriptsize QP\_CH\_NF}
        \label{fig:b}
    \end{subfigure}%
    \hfill
    \begin{subfigure}[b]{0.24\textwidth}
        \centering
        \includegraphics[width=\linewidth]{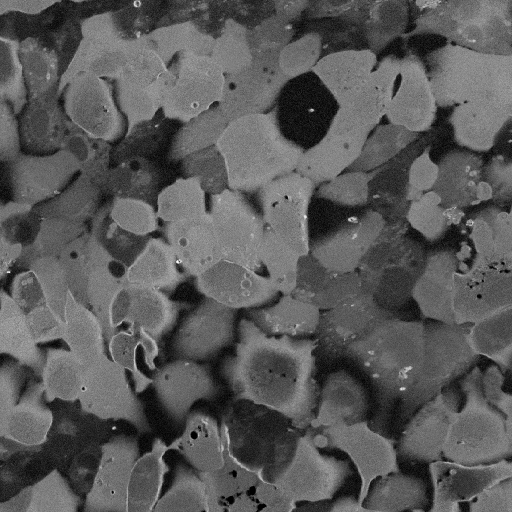}
        \caption{\scriptsize SDP\_GH\_CH}
        \label{fig:c}
    \end{subfigure}%
    \hfill
    \begin{subfigure}[b]{0.24\textwidth}
        \centering
        \includegraphics[width=\linewidth]{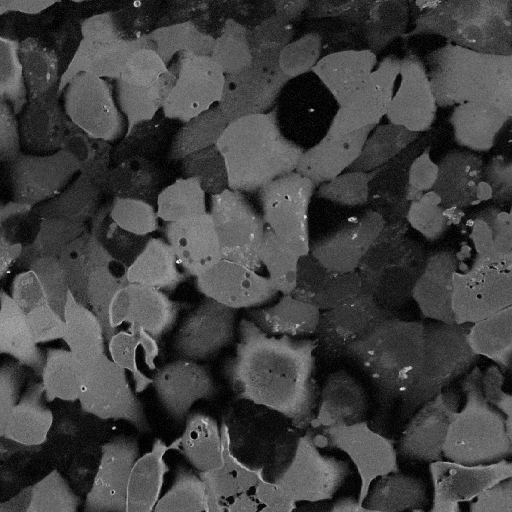}
        \caption{\scriptsize SDP\_GL\_CH\_BF}
        \label{fig:d}
    \end{subfigure}%

    \caption{Selected preprocessing and projection methods, reviewed with an expert experimentalist for cell annotation, including illustrative examples.}
    \label{fig:selected_prjs}
\end{figure}

\paragraph{Pixel-Level Agreement (Semantic segmentation)} At this level, we evaluated the semantic agreement between $GT_{n}$ and $GT_{o}$ using several metrics: Intersection over Union (IoU), which measures the extent of overlap between the two GTs, Background  Mismatch (BG $M$), which quantifies the percentage of pixels labelled as background in one GT but pertaining to foreground in the other. Results of these evaluations are presented in Table \ref{tab:bg_inst}.  

As illustrated in Table \ref{tab:bg_inst}, we observed notable variance in IoU scores across the dataset. To better understand this variability, we visually examined three representative cases from $GT_{o}$ and $GT_{n}$, one with the lowest IoU (Figure \ref{fig:selected_prjs}(a) and (d)), one with the highest (Figure \ref{fig:selected_prjs}(c) and (f)), and one approximating the metric's median (Figure \ref{fig:selected_prjs}(b) and (e)). Our analysis revealed that images taken at higher microscope magnifications (Figure \ref{fig:selected_prjs}(c) and (f)) yielded higher IoU scores between $GT_{o}$ and $GT_{n}$, indicating  alignment between the two GTs. This suggests that increased resolution may facilitate consistent segmentation.

\begin{table}[ht!]
        \centering
         \begin{adjustbox}{max width = \textwidth}
        \begin{tabular}{c|c|c|c|c|c |c |c }
        \hline \hline
            Metric &IoU & BG M$_{o}$ & BG M$_{n}$ & $GT_{o}$ CA &  $GT_{n}$ CA  & $GT_{o}$ CC &  $GT_{n}$ CC   \\
            \hline
            mean    & 86.08              & 6.86  & 5.15 & 4106 & 5452 & 47 & 74 \\ 
            
            p25     & 84.19              & 5.06  & 3.46 & 2354 & 3495 & 35 & 50\\
            median  & 86.43              & 6.56  & 4.72 & 2820 & 4238 & 52 & 75\\
            p75     & 88.69              & 8.31  & 6.06 & 4172 & 6174 & 61 & 91\\
            min     & 75.89              & 1.52   & 1.64 & 1372 & 2503 & 12  & 11\\
            max     & 92.40              & 15.13  & 14.36 & 21952 & 19286 & 84 & 171 \\
            \hline \hline
        \end{tabular}
        \end{adjustbox}
        \caption{Pixel-Level agreement comparison. BG $M_{n}$ shows percentage of pixel which are BG in $GT_{n}$ and foreground in $GT_{o}$.  BG $M_{o}$ shows the inverse.}
        \label{tab:bg_inst}
\end{table}

Furthermore, in these high-magnification cases, $GT_{n}$ provided clearer delineation of cellular boundaries and background regions, particularly at the image corners, where segmentation is typically more challenging. In the lowest IoU examples (Figure \ref{fig:selected_prjs}(a) \& (d)), $GT_{n}$ revealed more complete cell structures and extended boundaries, indicating improved visibility and segmentation accuracy. 

Additionally, BG $M_{o}$, which highlights instances where regions previously labelled as cells in $GT_{o}$ were reclassified as background in $GT_{n}$, was found to be higher than BG $M_{n}$, suggesting that  noise reduction or enhanced visibility improved the accurate delineation of true cell regions in $GT_{n}$. 

    \begin{figure}[ht!]
    \subfloat[Lowest IoU $GT_{o}$]{\includegraphics[width=0.32\textwidth]{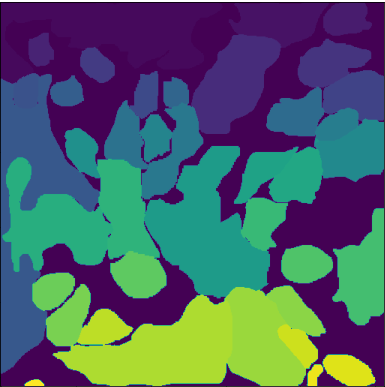}}
    \hfill
    \subfloat[Median IoU $GT_{o}$]{\includegraphics[width=0.32\textwidth]{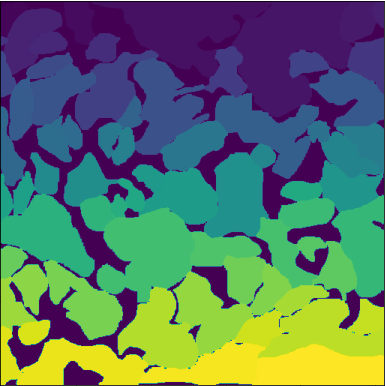}}
    \hfill
    \subfloat[Highest IoU $GT_{o}$]{\includegraphics[width=0.32\textwidth]{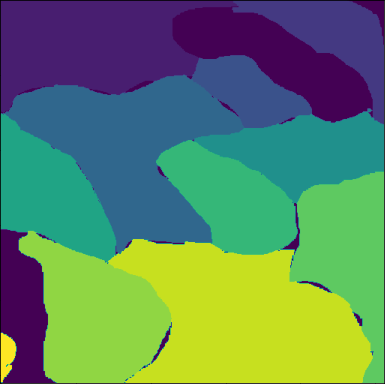}}
    
    \subfloat[Lowest IoU $GT_{n}$]
    {\includegraphics[width=0.32\textwidth]{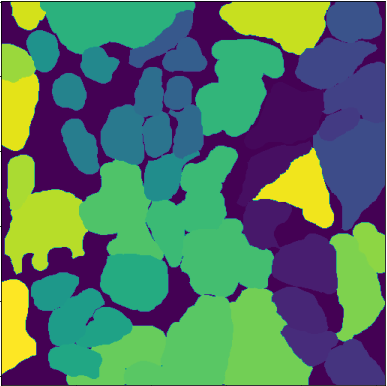}}
    \hfill
    \subfloat[Median IoU $GT_{n}$]{\includegraphics[width=0.32\textwidth]{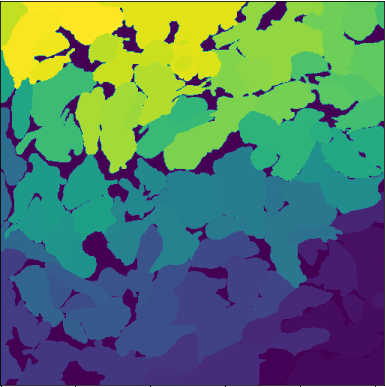}}
    \hfill
    \subfloat[Highest IoU $GT_{n}$]{\includegraphics[width=0.32\textwidth]{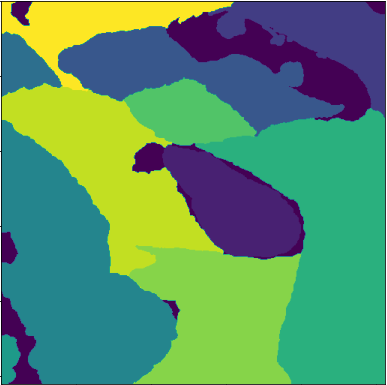}}
    \caption{Qualitative comparison between $GT_n$ and $GT_o$ as annotated by the experimentalist.}
    \label{fig:iou}
\end{figure}

\begin{figure} 
\subfloat[Cell area comparison \centering]{\includegraphics[width=0.48\textwidth]{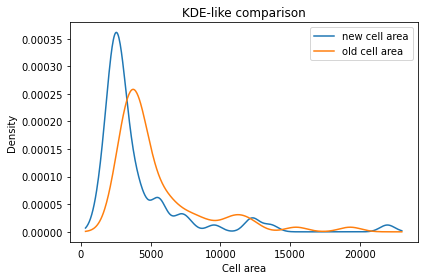}}
\hfill
\subfloat[Cell count comparison\centering]{\includegraphics[width=0.48\textwidth]{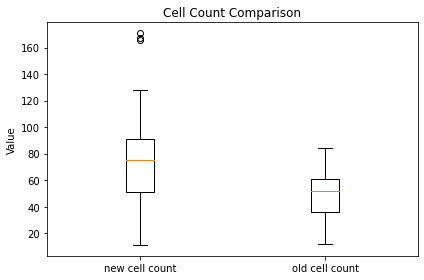}}
\hfill
\caption{Quantitative comparison $GT_n$ and $GT_o$. GT annotations were compared for the a) estimated cellular area and respective cellular density and for the b) overall number of cells detected within each image. }
\label{fig:box_kde}
\end{figure}

\paragraph{Cell-Level analysis (Instance segmentation)} This level focuses on instance-level comparisons. We examined metrics such as Cell Counts (CC), Cell Size (CS), and shape descriptors to assess structural differences between the GTs. The outcomes of this analysis are summarised in Table \ref{tab:bg_inst} and  visualized in Figure \ref{fig:box_kde}, using Kernel Density Estimation (KDE) and box-plots. 

The distribution of cell areas in $GT_{n}$ exhibits a pronounced peak at smaller sizes, indicating that the new segmentation and preprocessing pipeline tends to produce a greater number of small cells, with many cell areas clustered around lower values. In contrast, the distribution for $GT_{o}$ is broader, suggesting a wider range of cell sizes and less concentration around any particular value. This pattern implies that $GT_{n}$ yields more consistent and generally smaller cell segmentations, whereas $GT_{o}$ contains larger cells.

\section{Discussion}\label{Sec:Discussion}

Results achieved in this study show the potential of our proposed framework to overcome challenges associated with the analysis of temporally resolved video recordings of oscillatory fluorescence signals in live-cell FM experiments by fusing multi-temporal frames into higher quality 2D visualisation of these dynamic processes, in our case, $Ca^{2+}$ signalling. 

One of the key requirements for efficient image fusion is that the fused image must preserve all the quantitative information contained within the original video recording and contain more detailed information about the scene or object depicted than the isolated frames alone. Cell-level comparison of CA and CC in GT$_o$ and GT$_n$ annotations proved that images generated through the image fusion method presented in this study contain a higher number of cells, averaging a $44.23 \%$ increase in cell count (from a median of 52 cells identified in $GT_{o}$ to a median of 75 cells in $GT_{n}$), than images generated using conventional frames stacking and averaging methods (Table \ref{tab:bg_inst} and Figure \ref{fig:stats_box}b). These results suggest two main things: 1) our pipeline is more effective than previously employed methods \cite{caraffini2024towards} at combining data from multiple frames while reducing temporal inconsistencies across frames caused by fluctuating fluorescence signal intensity within cells. Therefore, projected images generally retain more complete information about the cellular network for downstream annotation (Figure \ref{fig:box_kde}); 2) efficient noise removal and contrast enhancement by our proposed approach improves the visibility of cellular boundaries, facilitating the separation of individual cells within dense and overlapping cellular regions characterised by cellular aggregation (Figure \ref{fig:iou}). This last assumption is further corroborated by the results achieved from the analysis of pixel-level agreement between GT$_o$ and GT$_n$. By comparing BG $M_o$ and BG $M_n$ (Table \ref{tab:bg_inst}), is therefore evident that a higher percentage of pixels identified as cells in GT$_o$ were changed to background pixels in GT$_n$ than vice-versa. This shift suggests that noise was either removed by our projection pipeline or that improved visibility and image sharpness, especially around cellular boundaries, enabled more accurate identification of true cell regions. Therefore, as shown in Figure \ref{fig:iou}a and \ref{fig:iou}d, large cells in $GT_{o}$ were often segmented into multiple smaller cells in $GT_{n}$, likely due to enhanced boundary detection. This is also suggested by the lower CA and higher cellular density recorded in $GT_{n}$ than $GT_{o}$ (Figure \ref{fig:box_kde}), where the lower amount of larger cells recorded in $GT_{o}$ suggests the under-segmentation of low-brightness cellular regions with unclear boundaries from unprocessed composite grayscale images. Overall, these observations support the hypothesis that the revised preprocessing pipeline produces more precise and biologically consistent segmentations.

The evaluation of perceived image quality using NR-IQA scores to determine the best preprocessing-projection combination pipeline leading to improved image quality yielded interesting yet inconclusive results (see Section \ref{sec:NR-IQAres}). First of all, while PIQE and NIQE scores were shown to largely fall within the range of values commonly observed  in NSS \cite{mittal2012making, venkatanath2015blind}, BRISQUE was found to behave in an unpredictable way when employed on our specific dataset, suggesting the inadequacy of the metric to provide insightful information about image quality when used on images radically different from the natural images on which the BRISQUE algorithm was trained, such as microscopy images. A similar behaviour for BRISQUE was found in \cite{dohmen2025similarity}. This study showed that, when employed for the evaluation of perceived quality in medical images, BRISQUE tends to assign high quality scores to images purposely distorted using light stripe artifacts, gaussian blur and ghosting \cite{dohmen2025similarity}, independently from the normalisation method use for image conversion to 8-bit format. However, as in our study (see Figure \ref{fig:scores_visual}), both BRISQUE and NIQE score were found to rightly assign the lowest quality values to heavily distorted images within the dataset.

Whilst some studies have shown the value of using PIQE for the evaluation of perceptive quality in medical images \cite{higashiyama2024investigation}, our results showed inconsistencies between images assigned HPQ by PIQE (Figure \ref{fig:Scores_vs_bio}a) and by the expert experimenter (Figure \ref{fig:Scores_vs_bio}b). Amongst the three metrics NIQE was proven to be the most appropriate to use for the identification of perceived high and low quality images within the HL-1 dataset (Figure \ref{fig:Scores_vs_bio}). Therefore, we found some concordance between the projection method (i.e., SDP) identified as best by NIQE and the methods selected by the experimenter for annotation (reported in \ref{fig:selected_prjs}), but no concordance across preprocessing methods employed (Figure \ref{fig:selected_prjs} and \ref{fig:best_averaged}). Based on these observations, despite in \cite{dohmen2025similarity}, the authors advise the readers against the use of NIQE for medical images, we instead suggest the potential use of NIQE to provide guidance for quality assessment in FM datasets. However, we advise to use this metric with caution and only as a support evaluation system to quantitatively corroborate the expert's perception, which we found to always outperform all NR-IQA metrics in the assessment of HPQ images within our dataset (Figure \ref{fig:Scores_vs_bio}).

\begin{figure}[ht!]

\begin{subfigure}{0.32\textwidth}
    \centering
    \includegraphics[width=\linewidth]{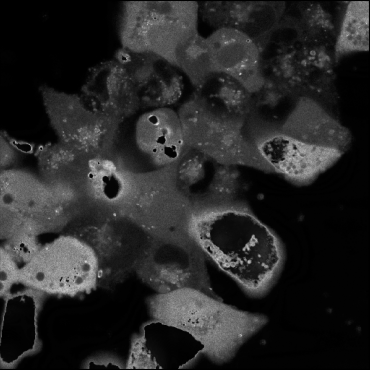} 
\end{subfigure}
\begin{subfigure}{0.32\textwidth}
    \centering
    \includegraphics[width=\linewidth]{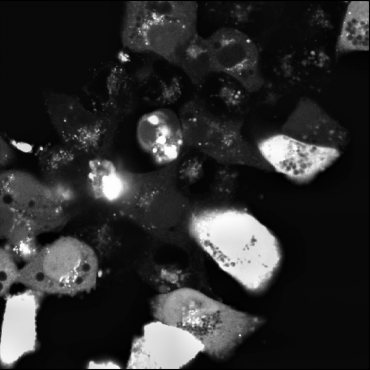} 
\end{subfigure}
\begin{subfigure}{0.32\textwidth}
    \centering
    \includegraphics[width=\linewidth]{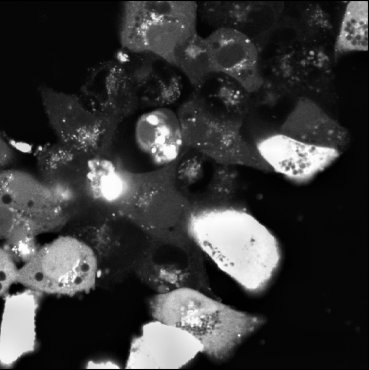}
\end{subfigure} \\[0.2em] 
\begin{subfigure}{0.32\textwidth}
    \centering
    \includegraphics[width=\linewidth]{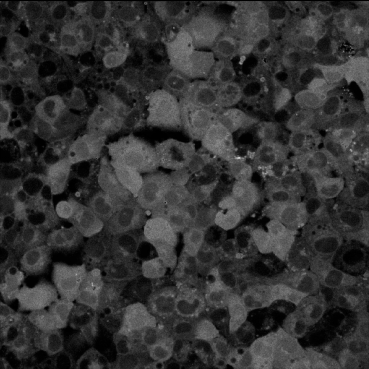}
    \caption{IQR}
\end{subfigure}
\begin{subfigure}{0.32\textwidth}
    \centering
    \includegraphics[width=\linewidth]{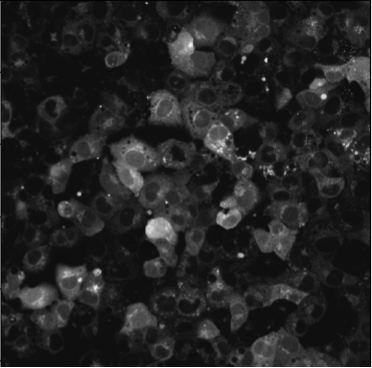}
    \caption{MDP}
\end{subfigure}
\begin{subfigure}{0.32\textwidth}
    \centering
    \includegraphics[width=\linewidth]{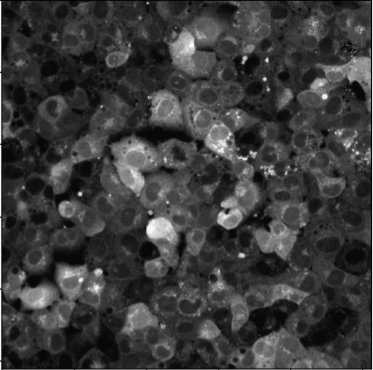}
    \caption{QP}
\end{subfigure}

\caption{Visual comparison of commonly used dispersion Projection Methods, (a) IQR and (b) MD, and (3) QP. Quantile-based projection tends to preserve border details more effectively than MD and IQR. Therefore, some cellular information is lost and some cells remain invisible in MDP compared to QP. It also better highlights borders and patterns of certain regions, such as brighter areas presumably corresponding to non-viable (dead) cells, better than IQR. providing clearer delineation in some regions. This justifies the use of QP instead of IQR and MDP in this study.}
\label{fig:compare_prjs}
\end{figure}

Amongst the six projection methods tested in the study, MIP, QP and SP methods were selected by the experimentalist as the methods yielding HPQ images (Figure \ref{fig:compare_prjs}). The preprocessing-projection combination methods selected were shown to generate images with high foreground-to-background contrast, which, together with high image brightness, facilitates the differentiation of cellular and background material in HL-1 images. These methods were also associated with an increased perceptive visibility and sharpness at cellular boundaries, making cell-to-cell connection more visible and facilitating the experimentalist in the recognition of cellular structures that characterise complex HL-1 networks for downstream GT annotation (as discussed in Section \ref{sec:GTcomp}). Interestingly, CH and GH preprocessing methods in combination in two out of four of these methods, associated with MP and SP projections, and again in their lower value version (i.e., GL and CL) with SP projections and CL alone with QP methods (as shown in Figure \ref{fig:selected_prjs}). Notably, combined CH and GH approaches, preceded or followed by filtering methods, MB, NF and BF amongst the most common, also appear in the best methods combinations identified by PIQE, NIQE and BRISQE scores (see Figure \ref{fig:best_averaged}). The value of using CLAHE with GF methods and NF techniques has been previously shown in medical imaging datasets \cite{Ajana@CLAHEgamma}, but never tested on FM datasets and in combination with projection methods for multi-temporal image fusion.

\subsection{Recommendations}\label{sec:Recommendations}
Based on the evidence provided in this study and the points raised in our discussion, we draw the following recommendations.

Using CLAHE and GC with high or low parameters (which may require tuning for specific datasets) and generally in combination yields good processing results, generating high quality images characterised by increased foreground-to-background contrast levels and sharpness at boundaries, which facilitate object recognition and segmentation for downstream analysis. 
 
Using noise filtering methods, especially MB, NF and BF after CLAHE and GH further enhances image quality by cleaning images from noise (e.g., salt and pepper noise) and reducing artifacts generation after fusion.

Employing MP, QP and SP z-projection methods in combination with the preprocessing techniques mentioned above yields optimal results for video-to-image processing of multi-temporal image stacks, generating fused images containing comprehensive and detailed information about the scene or object of interest.

Assessing image quality using NR-IQA metrics, specifically NIQE, can guide in assessing the quality of the obtained fused images, but quantitative analysis using this metric must always be corroborated with the experimentalist's expertise to decide which method provides the most useful results for annotation.

\section{conclusion and the way forward}\label{sec:conclusions}
Our study introduces an effective framework that enhances visual quality, fine-detail representation, and boundary recognition in FM, and proposes, for the first time, a z-projection-based methodology for fusing multi-temporal FM video frames of living tissue into comprehensive high-resolution images, thereby facilitating image segmentation and analysis.

We validate our methodology by using an exemplary dataset comprising real-time video datasets acquired from experiments on live HL-1 cardiac cell networks. Our results show that preprocessing frames using combined CH and GC methods, both with high and low parameters, followed by noise filtering and frame fusion using MIP, QP or SP z-projection methods significantly enhance image quality for downstream dataset annotation, improving objects and boundaries visibility and leading to an average increase in the amount of information retained and analysable within the fused image of around $44.23 \%$. This large experimentation phase enabled for general recommendations on the combined use of computer vision techniques in this domain, and led us to assess suitability of several NR-IQA metrics for the assessment of perceived image quality in microscopy datasets other than for natural images.Although these metrics do not seem to work well with microscopy imaging, NIQE—supported by expert visual inspection—may still guide image quality assessment for our dataset.

Future work should apply our pipeline to larger imaging datasets, including cellular images, and optimise new NR-IQA metrics for microscopy image quality. Our pipeline may be broadly applicable to diverse imaging types and modalities that fuse multi-temporal image stacks from live videos into high-quality 2D images, facilitating downstream annotation and segmentation. In addition, the pipeline might benefit by focusing on enhancing z-projection through alternative approaches such as the Karhunen–Loève Transform, Principal Component Analysis, incorporating area-averaged peak detection for projection, and systematically evaluating the effect of varying the number of input videos on projection accuracy.

\section{Appendices}\label{sec:appendices}

\section*{CRediT authorship contribution statement}

Hassan Eshkiki: Software, Writing – original draft, Writing – review \& editing, Data curation, Conceptualisation, Investigation, Methodology, Supervision, Validation;
Sarah Costa: Formal analysis, Investigation, Writing – original draft, Writing – review \& editing, Data curation, Visualisation, Validation; 
Mostafa Mohammadpour: Methodology
Farinaz Tanhaei: Software, Investigation, Validation; 
Christopher H. George: Writing – review \& editing, Resources, Funding acquisition, Supervision;
Fabio Caraffini: Writing – original draft, Writing – review \& editing, Resources, Investigation, Funding acquisition, Supervision, Validation. 

\section*{Data availability}

We provide sample data through a Zenodo repository \cite{bib:ProjectionZenodoRepo}, which contains supporting materials, including processing code. All raw image data used in this study can be made available upon reasonable request.

\section*{Declaration of competing interest}
The authors declare that they have no known competing financial interests or personal relationships that could have appeared to influence the work reported in this article. 

\section*{Acknowledgment}

This research is supported by the Morgan Advanced Studies Institute, Wales, UK; the National Cardiovascular Research Network (funded by Health and Care Research Wales); and the British Heart Foundation (Reference number: RG/15/6/31436). For the purpose of Open Access, the authors have applied a CC BY license to any Author Accepted Manuscript (AAM) version arising from this submission.

\end{document}